%% file: main_iccv.tex
\documentclass[10pt,twocolumn,letterpaper]{article}

\usepackage[pagenumbers]{iccv}

\input{preamble}

\definecolor{iccvblue}{rgb}{0.21,0.49,0.74}
\usepackage[pagebackref,breaklinks,colorlinks,allcolors=iccvblue]{hyperref}
\usepackage{multirow}
\usepackage{colortbl}
\usepackage[most]{tcolorbox}
\usepackage[title]{appendix}
\usepackage{fontawesome}

\newcommand{\appref}[1]{Appendix~\ref{#1}}
\newcommand{\unseen}{\red{\textbf{Unseen Tasks}~\faLowVision :}}

\def\confName{ICCV}
\def\confYear{2025}

\title{\ourmethod: A Universal Image Generation Framework \\ 
via Visual In-Context Learning}

\author{Zhong-Yu Li\textsuperscript{1,4}\footnotemark[1] \quad Ruoyi Du\textsuperscript{2,4}\footnotemark[1] \quad Juncheng Yan\textsuperscript{3,4} \quad Le Zhuo\textsuperscript{4} \quad Qilong Wu\textsuperscript{4} \\ Zhen Li\textsuperscript{5}\footnotemark[2]  \quad
Peng Gao\textsuperscript{4} \quad Zhanyu Ma\textsuperscript{2} 
\quad Ming-Ming Cheng\textsuperscript{1}\footnotemark[2] \\ 
\textsuperscript{1}VCIP, CS, Nankai University 
\quad
\textsuperscript{2}Beijing University of Posts and Telecommunications \\
\textsuperscript{3}Tsinghua University
\quad
\textsuperscript{4}Shanghai AI Laboratory
\quad
\textsuperscript{5}The Chinese University of Hong Kong \\
\faLink~Project page: \href{https://visualcloze.github.io}{\textcolor{red}{https://visualcloze.github.io}}
}

\begin{document}

\twocolumn[{
    \renewcommand\twocolumn[1][]{#1}
    \maketitle
    \begin{center}
        \centering
        \vspace{-20pt}
        \includegraphics[width=0.98\linewidth]{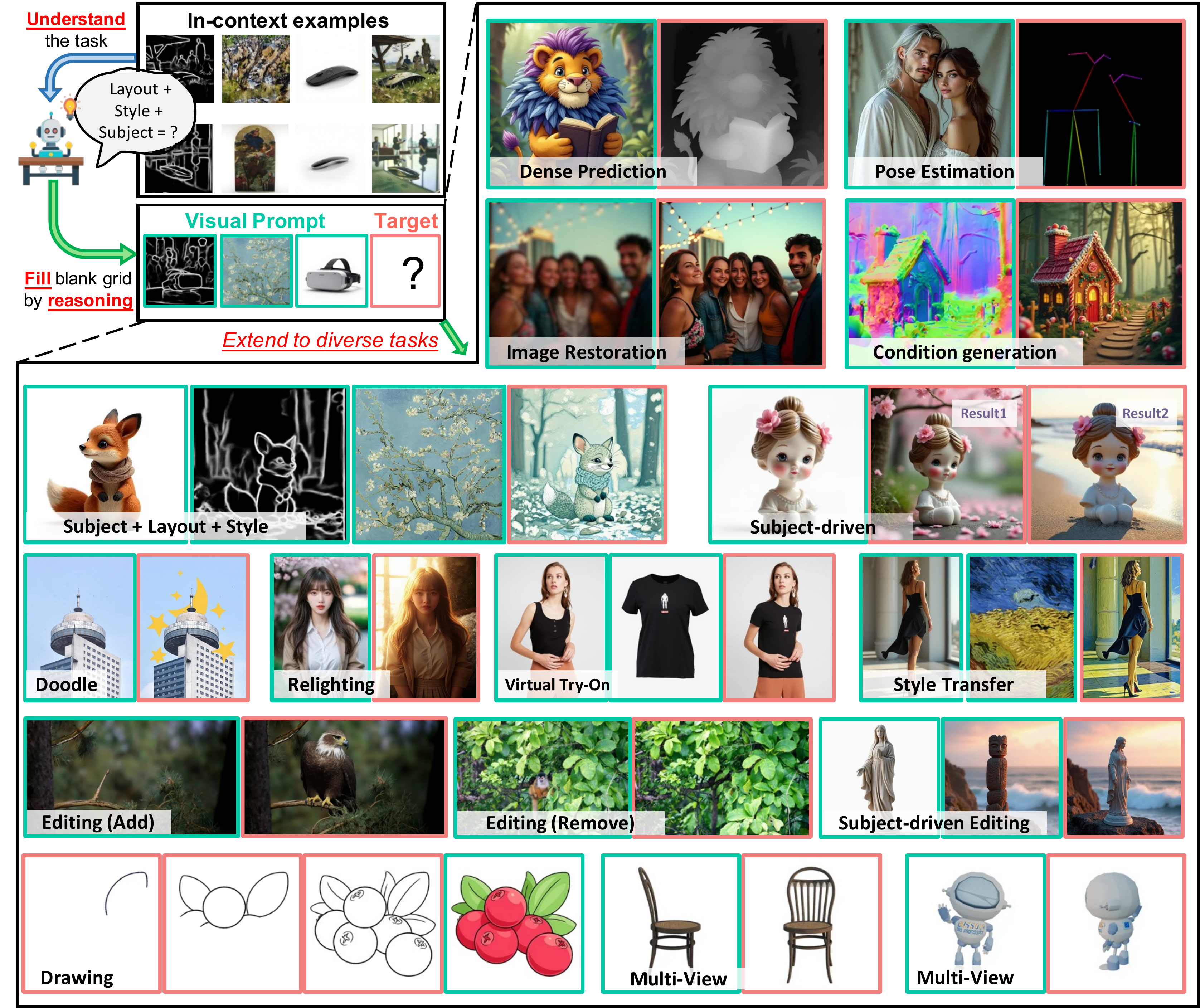}
        \vspace{-8pt}
        \captionsetup{type=figure}
        \caption{The top left illustrates our universal image generation 
        framework based on visual in-context learning. 
        Given one query of a specific task, 
        the generative model learns the task by observing a few in-context examples presented as demonstrations. 
        For each task, \red{the generation result is indicated by a red box}.
        }
        \label{fig:emergence}
    \end{center}
}]

\twocolumn[{
    \renewcommand\twocolumn[1][]{#1}
    \begin{center}
        \centering
        \includegraphics[width=1\linewidth]{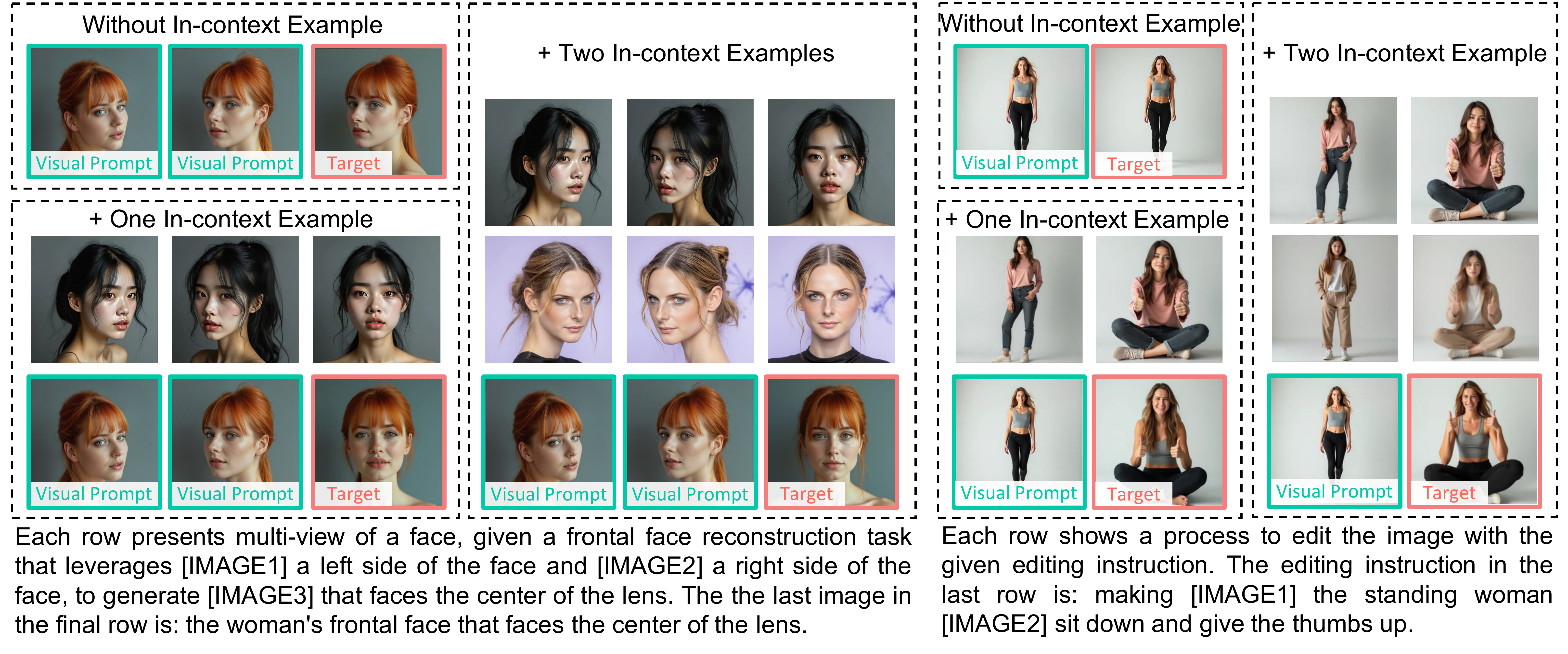}
        \captionsetup{type=figure}
        \vspace{-25pt}
        \caption{\unseen~Generalizing to {{tasks unseen during training}} via in-context learning. 
        More in-context examples lead to more accurate results.}
        \label{fig:unseen}
    \end{center}
}]

\let\thefootnote\relax\footnotetext{$^*$ Equal contribution\hspace{3pt} \hspace{5pt}$^\dagger$ Corresponding author\hspace{5pt}
}

\input{sec/0_abstract}    
\input{sec/1_intro}
\input{sec/2_related_work}
\input{sec/3_dataset}
\input{sec/4_method}
\input{sec/5_experiment}
\input{sec/6_limitation}
\input{sec/7_conclusion}
{
    \small
    \bibliographystyle{ieeenat_fullname}
    \bibliography{main}
}

\input{sec/8_appendix}

\end{document}

%% file: preamble.tex
\usepackage{overpic}
\newcommand{\red}[1]{{\color{red}#1}}

\newcommand{\ourdataset}{Graph200K}
\newcommand{\ourmethod}{VisualCloze}
\newcommand{\myPara}[1]{\vspace{.15in} \noindent\textbf{#1}}
\newcommand{\figref}[1]{Fig.~\ref{#1}}
\newcommand{\tabref}[1]{Tab.~\ref{#1}}
\newcommand{\secref}[1]{Sec.~\ref{#1}}
\renewcommand{\eqref}[1]{Equ.~(\ref{#1})}

%% file: sec/0_abstract.tex
\begin{abstract}
Recent progress in diffusion models significantly advances various image generation tasks.
However, the current mainstream approach remains focused on building task-specific models, which have limited efficiency when supporting a wide range of different needs.
While universal models attempt to address this limitation, 
they face critical challenges, including generalizable task instruction, appropriate task distributions, and unified architectural design. 
{To tackle these challenges, 
we propose \ourmethod, a universal image generation framework, 
which supports a wide range of \textbf{in-domain tasks}, 
generalization to \textbf{unseen} ones, 
\textbf{unseen unification} of multiple tasks, and \textbf{reverse} generation.}
Unlike existing methods that rely on language-based task instruction, 
leading to task ambiguity and weak generalization, 
we integrate visual in-context learning, 
allowing models to identify tasks from visual demonstrations.
Meanwhile, 
the inherent sparsity of visual task distributions 
hampers the learning of transferable knowledge across tasks. 
To this end, we introduce \ourdataset, 
a graph-structured dataset that establishes various interrelated tasks, 
enhancing task density and transferable knowledge. 
Furthermore, we uncover that our unified image generation formulation shared a consistent objective with image infilling, 
enabling us to leverage the strong generative priors of 
pre-trained infilling models without modifying the architectures.
\end{abstract}
\vspace{-20pt}

%% file: sec/1_intro.tex
\section{Introduction}
\label{sec:intro}

\begin{figure*}[t]
	\centering
	\begin{overpic}[width=1.0\linewidth]{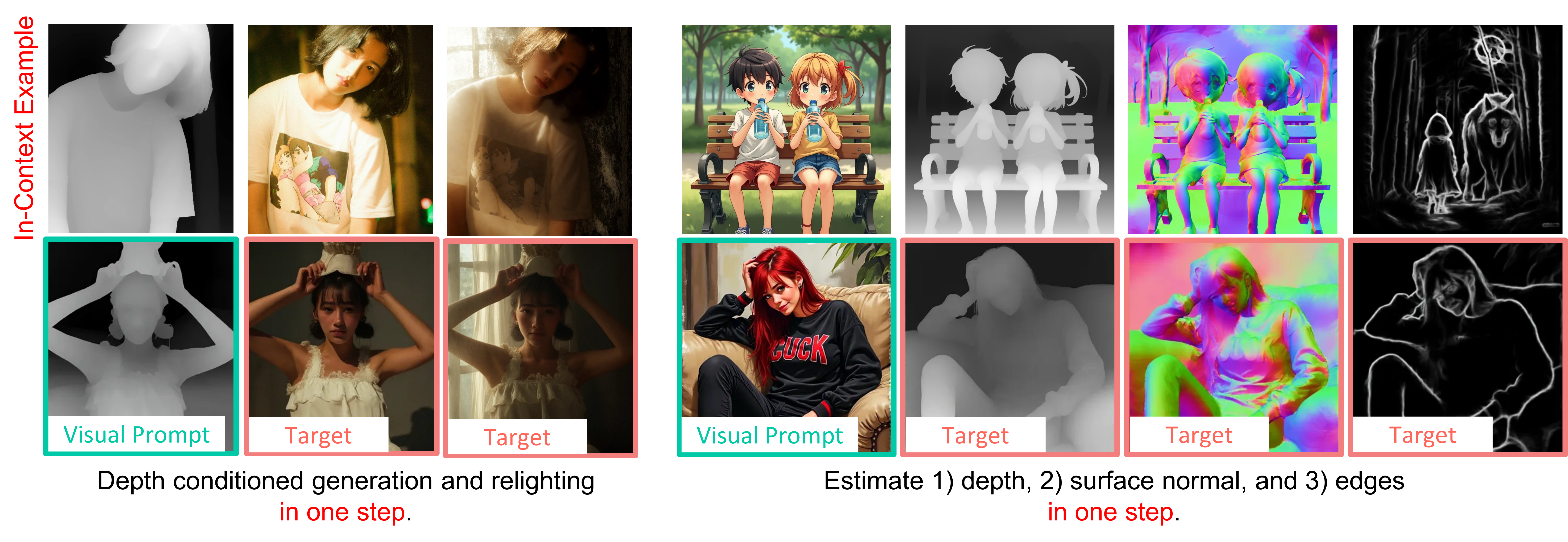}
	\end{overpic}
    \vspace{-25pt}
	\caption{\unseen~Leveraging in-context learning to unify multiple seen tasks into {a single-step unseen task}. 
    Left: Unifying the [Depth to Image] and [Relighting] task into a single [Depth to Images with Various Lighting] task.
    Right: Unifying multiple dense prediction tasks into a joint prediction task. 
    Results without visual context can be found in the appendix.
    }
    \vspace{-10pt}
    \label{fig:combination}
\end{figure*}

Recent advancements in image generation, 
propelled by the progress of diffusion models~\cite{esser2024scalingrectifiedflowtransformers,zhuo2024luminanext,flux2024}, 
have led to a wide range of applications, 
including image editing~\cite{wei2024omniedit}, 
style transfer~\cite{wang2024instantstyle, Zhang_2023_inst}, 
virtual try-on~\cite{choi2021viton, chong2025catvton}, 
and personalized generation~\cite{li2023photomaker, ruiz2023dreambooth}, 
among others. 
However, these tasks typically require task-specific models, 
which limit efficiency and scalability for real-world applications.
In recent years, 
there has been growing interest in universal generative models~\cite{lin2024pixwizard,lhhuang2024iclora,mao2025ace++}, 
aiming to handle diverse image generation tasks, even unseen ones, within a single unified framework.
Despite significant progress, some critical issues remain to be addressed, such as (1) distinguishable and generalizable task instruction, (2) comprehensive task coverage during training, and (3) a unified model architecture. 

An ideal task instruction is crucial for guiding the model to process the desired task effectively.
Existing methods primarily rely on language instructions~\cite{lhhuang2024iclora,mao2025ace++} or task-specific 
tokens~\cite{lin2024pixwizard} to distinguish the task to be performed. 
However, the complexity of visual tasks and the inherent gap between vision 
and language modalities make it hard for the model to understand language-only task descriptions,
which leads to task confusion~\cite{lin2024pixwizard} and 
hinders generalization on unseen tasks~\cite{xiao2024omnigen,le2024diffusiongenerate}. 
Moreover, pre-learned task-specific tokens constrain the model only to handle seen tasks.
In contrast, 
large language models (LLMs) have successfully achieved unified multi-task modeling, 
partially due to the rise of in-context learning~\cite{brown2020language}, 
which allows models to adapt various tasks 
using only a few demonstrations. 
We aim to replicate the concept of in-context learning in the pure visual modality, 
{where the model learns the desired task directly from a few visual examples as task demonstrations, as shown in \figref{fig:emergence} (Left Top).}
In this setting, in-context learning shows strong potential for universal image generation.
We summarize four key findings:
(1) it supports various in-domain tasks with reduced task ambiguity~(\figref{fig:emergence});
(2) it generalizes to unseen tasks~(\figref{fig:unseen}, \figref{fig:multi_subject}); 
(3) {as an unseen strategy for task unification, it can integrate multiple sub-tasks into a single step and generate intermediate results~(\figref{fig:combination})};
(4) it enables reverse generation, \ie, inferring a set of conditions from a given target~(\figref{fig:reverse}).
While prior works~\cite{xiao2024omnigen,bar2022visual,zhang2023what,balazevic2023towards,Wang_2023_CVPR,liu2023unifying,alayrac2022flamingo} 
have also explored in-context learning in vision, they are largely constrained to specific domains 
(such as dense prediction or style transfer~\cite{Wang_2023_ICCV,zhu2024unleashing}), 
or simplified generation settings involving only one condition and one target image~\cite{liu2023unifying,sun2023imagebrush}. 

From the perspective of task distribution, 
visual tasks are inherently sparse compared to those in natural language processing 
because task-specific datasets~\cite{xiao2024omnigen,zhou2017scene} for different tasks have minimal overlap~\cite{Yun_2023_ICCV,Kokkinos_2017_CVPR,Ghiasi_2021_ICCV}. 
Such sparse task learning isolates the knowledge of each task and 
limits the model from 
learning shared features across tasks. 
Moreover, the weak correlations between tasks hinder knowledge transfer and adaptability to new tasks. 
However, 
existing works in multi-task learning~\cite{ruder2017overview,Kendall_2018_CVPR,fifty2021efficiently,chen2018gradnorm} 
have verified the benefits of overlapping knowledge across related tasks. 
To alleviate the sparsity of visual tasks, 
we introduce a graph-structured dataset, \ourdataset, 
where each image is associated with annotations spanning five meta-tasks, 
\ie, conditional generation~\cite{zhang2023adding}, 
IP preservation~\cite{ye2023ip-adapter}, style transfer~\cite{Zhang_2023_inst}, 
image editing~\cite{wei2024omniedit}, and restoration~\cite{yu2024scaling}. 
By combining different conditions, 
we train the model with a variety of tasks that overlap with each other. 
Given this highly overlapping and compact task space, 
our dataset significantly increases task density, 
allowing the model to learn shared and transferable {knowledge} more effectively.

For the architecture design, it is essential to 
1) accommodate flexible task formats~\cite{lhhuang2024iclora,xiao2024omnigen,le2024diffusiongenerate}, ensuring seamless in-context learning, 
and 2) remain compatible with state-of-the-art models~\cite{zhuo2024luminanext,flux2024} to fully leverage their strong generative priors. 
In this work, 
we find that the state-of-the-art image infilling model~\cite{flux2024} 
has a consistent objective with our in-context learning based universal generative formulation. 
Specifically, 
we concatenate all input and output images together, 
where the objective of a task is to fill the output area. 
This alignment enables us to build our model upon advanced general-purpose infilling models 
without additional modifications, achieving powerful universal generation capabilities with minimal data and training costs. 

In this work, 
we propose a universal image generation framework, 
\ourmethod, which fine-tunes FLUX.1-Fill-dev~\cite{flux2024} with 
interrelated tasks sampled from \ourdataset~to 
learn transferable knowledge and support visual in-context learning. 
As the number of in-context examples increases, 
we observe enhanced performances and reduced task confusion, 
enabling the model to support a broad spectrum of in-domain tasks, 
including conditional generation, image restoration, editing, style transfer, IP-preservation, and their combinations. 
On unseen tasks, 
the model also shows a certain degree of generalization ability, 
as shown in \figref{fig:unseen}. 
In summary, 
our main contributions are as follows:
\begin{itemize}
    \item We propose an in-context learning based universal image generation framework that supports a wide range of in-domain tasks 
    and exhibits generalization to unseen ones.
    \item We design a graph-structured dataset, \ourdataset, which constructs a compact task space, enabling flexible online task sampling and 
    promoting the models to learn shared and transferable {knowledge} across tasks. 
    \item {Our unified image generation formulation shares a 
    consistent objective with the state-of-the-art infilling model}, 
    enabling exceptional performance through minimal tuning 
    without modifying the structure.
\end{itemize}

%% file: sec/2_related_work.tex
\section{Related Work}
\label{sec:related_work}

\subsection{Image Generation}
Recent advances in text-to-image generation have achieved remarkable 
performance, largely driven by the development of autoregressive models~\cite{sun2024autoregressive,yu2022scaling,liu2024lumina-mgpt} and 
diffusion models~\cite{ddpm,gao2024mdtv2maskeddiffusiontransformer,Peebles_2023_ICCV,dhariwal2021diffusion,Rombach_2022_CVPR,esser2024scalingrectifiedflowtransformers,albergo2023building,lipman2023flow,liu2022flow}. 
Among these, rectified flow transformers~\cite{flux2024,esser2024scalingrectifiedflowtransformers,gao2024lumina,zhuo2024luminanext} 
have shown great training efficiency and overall performance. 
Building on these foundational models, 
diverse applications have emerged, 
such as conditional generation~\cite{zhang2023adding}, 
style transfer~\cite{wang2024instantstyle}, and 
personalized generation~\cite{li2023photomaker}. 
More recently, universal models that address various tasks~\cite{zhao2025diceptiongeneralistdiffusionmodel,mao2025ace++,le2024diffusiongenerate} have been explored. 
For example,
unified models like OmniGen~\cite{xiao2024omnigen} leverage large vision language models to
consolidate multiple tasks into a single framework.
Similarly, UniReal~\cite{chen2024UniReal} unifies image generation tasks as discontinuous video generation.
However, 
they still face issues such as 
over-reliance on language instructions, 
isolation and sparsity of visual tasks, 
and architecture design accommodating flexible task formats. 
To address these issues, 
we propose a universal image generation framework that unifies generation tasks as image infilling. 
Through visual in-context learning and our \ourdataset~dataset that constructs a denser task space to learn transferable {knowledge}, 
our method alleviates ambiguity to support a diverse set of 
in-domain tasks and generalizes to tasks unseen during training. 

\subsection{Visual In-context Learning}
Along with the emergence of large language models, such as GPT-3~\cite{brown2020language}, 
in-context learning~\cite{dong2024surveyincontextlearning} has been an effective approach 
to allow the language model to 
understand and perform complex tasks 
given a few demonstrations. 
Early works~\cite{imageanalogies,hertzmann2001algorithms} 
in vision modality propose 
image analogies to 
create an image filter from examples automatically. 
In recent years, 
leveraging 
inpainting model~\cite{bar2022visual,zhang2023what,balazevic2023towards}, 
masked image modeling~\cite{Wang_2023_ICCV,Wang_2023_CVPR,liu2023unifying}, 
or vision-language model~\cite{alayrac2022flamingo,zhou2024visualincontextlearninglarge}, 
visual in-context learning is proposed to handle more tasks.
However, 
they mainly focus on 
dense prediction~\cite{zhu2024unleashing,Sheng_2024_CVPR,sun2023exploringeffectivefactorsimproving} or 
visual understanding~\cite{wang2024visincontext}. 
{OmniGen~\cite{xiao2024omnigen} also leverages in-context learning to 
generalize to unseen domains, 
\eg, segmenting unseen concepts when the model has learned the segmentation task during training. 
However, 
it mainly focuses on simple tasks of dense prediction, and 
the gap between the unseen and training domains is still limited.} 
Some recent works~\cite{liu2023unifying,sun2023imagebrush,wang2023incontext,lai2024unleashing} extend visual in-context learning 
to image generation, 
but they are still limited by simple tasks such as conditional generation and dense prediction. 
Moreover, 
the sparsity of visual tasks 
makes it difficult for models to learn transferable and overlapping knowledge across tasks, 
limiting the generation ability of in-context learning. 
In contrast, 
we introduce a graph-structured dataset that supports interrelated tasks and thus constructs a more dense task space, 
promoting the model to learn shared and transferable knowledge and enhance its adaptability. 

%% file: sec/3_dataset.tex
\section{Dataset}
\label{sec:dataset}

\begin{figure}[t]
	\centering
	\begin{overpic}[width=1.0\linewidth]{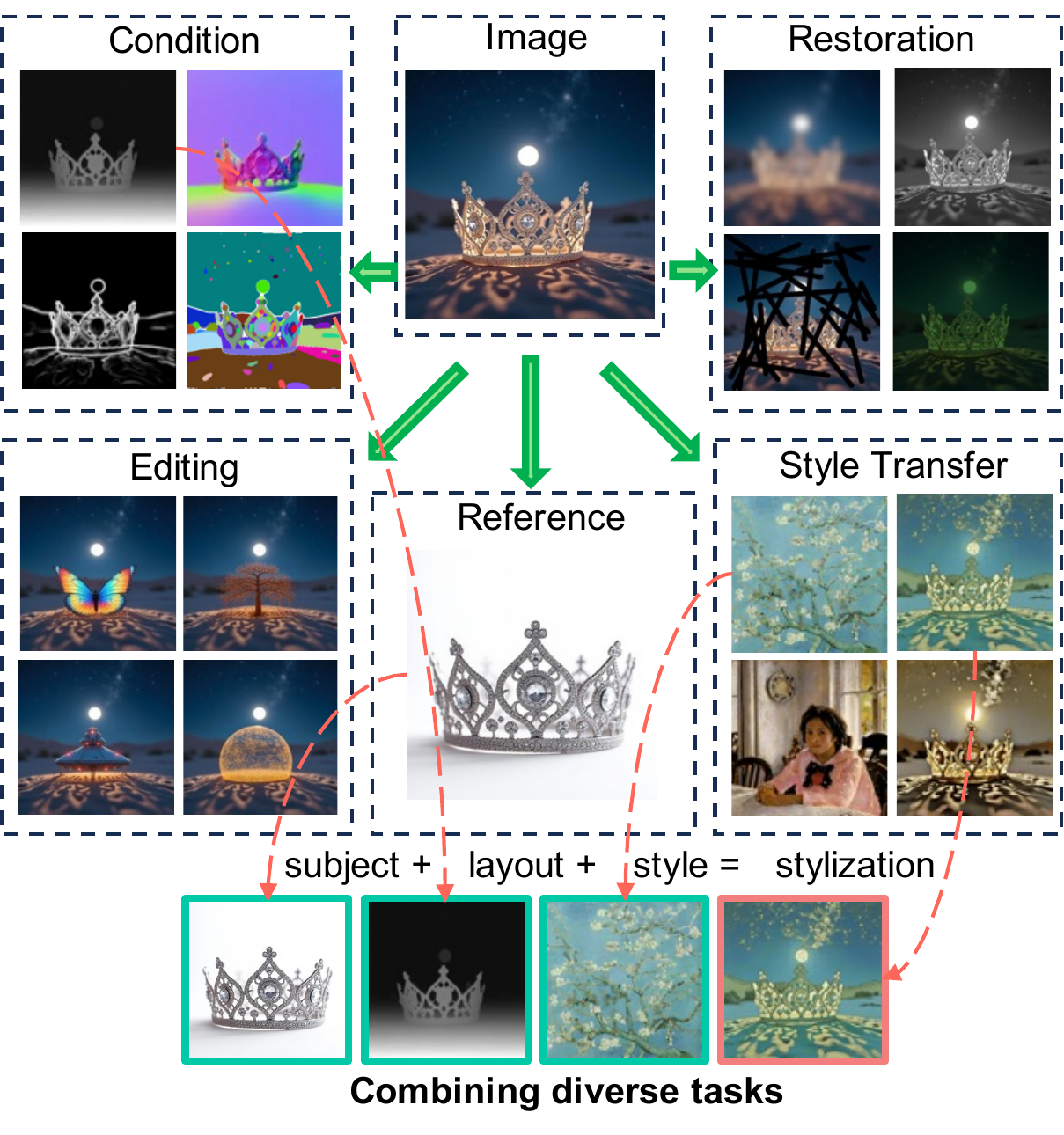}
	\end{overpic}
	\caption{Illustration of the proposed \ourdataset~dataset. 
        Each image is annotated for five meta-tasks, \ie, 
        conditional generation, image restoration, image editing, 
        IP preservation, and style transfer. 
        Using these tasks, 
        we can combine a wide range of complex tasks, such as the 
        bottom of the figure.
    }
    \label{fig:dataset}
\end{figure}

Recent works~\cite{mao2025ace++,lhhuang2024groupdiffusion,xiao2024omnigen} 
have made great progress in unified image generation. 
However, their generalization to unseen tasks 
remains highly limited. 
We partially attribute this issue to the sparsity and isolation of visual tasks, 
hindering the model from learning shared features across tasks and handling unseen ones. 
Moreover, weak correlations between tasks further hinder knowledge transfer, 
restricting the adaptability of models.
Therefore, increasing task density or strengthening task inter-relations helps improve the generalization ability of models via a compact task distribution.
{In this paper, we take the Subject200K~\cite{tan2024ominicontrol} dataset as a starting point 
and construct our \ourdataset~dataset by augmenting each image with 49 types of annotations spanning five meta-tasks. 
This enriched annotation space enables flexible construction of a wide range of related tasks 
by sampling and combining arbitrary subsets of annotations across different meta-tasks, 
as illustrated in \figref{fig:dataset}.}

\subsection{Graph-Structured Multi-Task Dataset}
\label{sec:grapth_structure}
In natural language processing, 
tasks overlap significantly, 
facilitating strong cross-task learning ability. 
In contrast, visual tasks are inherently distinct, posing challenges for vision models to achieve similar generalization ability via instruction tuning.
To ease this issue, 
we introduce a \emph{Graph-Structured Multi-Task Dataset}. 
As illustrated in \figref{fig:dataset} (a), 
given a text-to-image dataset, 
each image is treated as the central node of a graph, 
around 
which diverse task annotations are constructed, 
including those for various spatial conditions, 
degradations, 
image editing results, 
reference image for IP-preservation, 
and style transfer with various reference styles. 
The construction process for each task pair is detailed in the next section.

As shown in \figref{fig:dataset}, 
each task annotation forms a bidirectional edge with the image. 
Thus, the graph is strongly connected, 
which means that for any two nodes, 
bidirectional paths exist between them. 
In other words, 
a generation task can be formulated as a path within the graph. 
The nodes along a path (except the end node) serve as condition images, 
which is analogous to the question in instruction fine-tuning, 
while the target image~(the end node) plays the role of the answer.  
Specifically, 
there are {49} types of nodes in our \ourdataset, and 
we sample up to {134} highly overlapping tasks, 
making the model learn more compact and shared representations across tasks. 
Moreover, it enriches the diversity and flexibility of our instruction fine-tuning data. 
For example, 
the path ${\rm reference}\rightarrow{\rm editing}\rightarrow{\rm image}$ corresponds to 
the task of image editing with reference, 
as shown in \figref{fig:dataset} bottom.

\subsection{Dataset Construction}
For convenience, 
we inherit subject-driven data from the Subjects200K~\cite{tan2024ominicontrol}. 
Additionally, {32} different degradations are applied online to the images to 
acquire restoration data.
We summarize the data construction methods in this section for the remaining three tasks.

\myPara{Conditional generation.} 
Each image is paired with {12} distinct conditions 
generated by specialized models, 
including canny edges~\cite{canny1986computational}, HED edges~\cite{xie15hed}, Hough lines~\cite{gu2021realtime}, 
semantic segmentation maps~\cite{uniformer}, depth maps~\cite{depth_anything_v2}, 
shape normal maps~\cite{xu2023unifying}, and human keypoints~\cite{openpose}, 
following ControlNet~\cite{zhang2023adding}.
This work extends the conditions by 
incorporating SAM2~\cite{ravi2024sam2} masks, foreground segmentation, and open-world boxes and masks. 
The foreground segmentation, derived from the RMBG~\cite{BiRefNet}, supports 
diverse tasks such as inpainting and foreground extraction. 
Open-world bounding boxes are generated through the grounding caption capability of Qwen2-VL~\cite{Qwen2VL}, 
which are processed using SAM2~\cite{ravi2024sam2} to produce corresponding masks.

\myPara{Style transfer.}
We transfer the style of images according to reference 
in both semantic-variant and semantic-invariant settings.
Specifically, the semantic-invariant transfer adopts 
InstantStyle~\cite{wang2024instantstyle} to 
preserve the semantic content, 
while the semantic-variant transfer relies on FLUX.1-Redux-dev~\cite{flux2024}, 
using the style embeddings and depth as conditions. 
For each image, we randomly generate five stylized versions. 
Mixing the two tasks pushes the model to follow the in-context examples better to avoid ambiguity.

\myPara{Image editing.}
We design two types of editing tasks, 
including background-variant and background-invariant editing. 
The background-invariant editing begins with 
localizing the subjects. 
Then, we leverage a large vision-language model, 
Qwen2-VL~\cite{Qwen2VL}, 
to modify the image caption 
with a new object that replaces the original subject.
The image, with the subject masked, 
is subsequently processed by the FLUX.1-Fill-dev~\cite{flux2024} inpainting model to 
integrate the alternative object into the masked region. 
The above operation is repeated five times to enrich the dataset. 
For background-variant editing, 
the difference lies in 
the last step, 
which utilizes FLUX.1-Redux-dev~\cite{flux2024} 
with depth as the condition 
and the modified caption as the text prompt.

\subsection{Other Data}
{To further expand the range of tasks and 
enhance the generalization ability of models, 
we incorporate several open-source datasets during training, including 
VITON-HD~\cite{choi2021viton} for virtual try-on and 
PhotoDoodle~\cite{huang2025photodoodlelearningartisticimage} 
for artistic image editing. 
For image editing tasks,
we also extend the dataset with OmniEdit~\cite{wei2024omniedit}. 
Specifically, 
two sub-tasks, \ie, object addition and removal, are used for training. 
The other editing tasks, 
such as attribute modification and environment change, 
are treated as unseen tasks to assess the generalization ability of the trained model. 
Furthermore, we leverage a portion of 
high-quality internal data, 
covering tasks of the drawing process~\cite{paintsundo} and multi-view generation~\cite{huang2024mvadapter}.} 

%% file: sec/4_method.tex
\section{Method}
\label{sec:method}

This paper identifies the core challenges in building a universal image generation model, 
including the need for a clearly defined and generalizable task formulation,  visual task sparsity, and the lack of a unified framework for multi-task learning. 
In the previous section, we addressed the issue of task sparsity by constructing the compact Graph200K dataset. 
\secref{sec:incontext} introduces visual in-context learning 
as the ideal paradigm for universal task formulation. 
Afterward, \secref{sec:unified_framework} considers the 
image infilling model a unified multi-task framework, 
achieving strong generalization capabilities with minimal cost.

\subsection{Visual In-context Learning} 
\label{sec:incontext}
Language instructions are usually used to specify the generation definition to handle multiple visual generation tasks with a single generative model. 
However, due to the gap between vision and language, 
the text comprehension ability 
of image generation models remains limited. 
This issue leads to task confusion~\cite{lin2024pixwizard} in existing universal generative models and weak generalization to unseen tasks.
Inspired by the success of 
few-shot learning on large language models~\cite{brown2020language}, 
we recognize that visual context 
may serve as a more friendly 
task 
instruction 
for visual generative models, 
given their superior visual understanding capabilities.

Therefore, in this paper, we re-propose visual in-context learning to build a universal and generalizable image generation system.
{For the sake of description}, 
here we {assume} the image input-output of arbitrary conditional generation task as a query consisting of $L-1$ condition images and a blank target $\varnothing$ to be completed by the model, \emph{i.e.}, ${X}={\rm concat}(\{x_1, \dots, x_{L-1}, \varnothing\})$. 
In \secref{sec:qualitative}, we demonstrate that 
our method can be extended to more general scenarios, 
where it can generate images at arbitrary positions and in any quantity rather than just the single image at the end of the query.
{During training, we randomly provide up to $C$ in-context examples, 
each containing $L$ images as the query.} 
This strategy 
ensures the generalization ability of models across different numbers of in-context examples. 
In our experiments, 
we show that providing in-context examples as task demonstrations not only 
helps alleviate task confusion and boost model performance across in-domain tasks~\cite{lin2024pixwizard}, 
but also enhances the generalization ability on unseen tasks. 

\begin{figure}[t]
	\centering
	\begin{overpic}[width=1.0\linewidth]{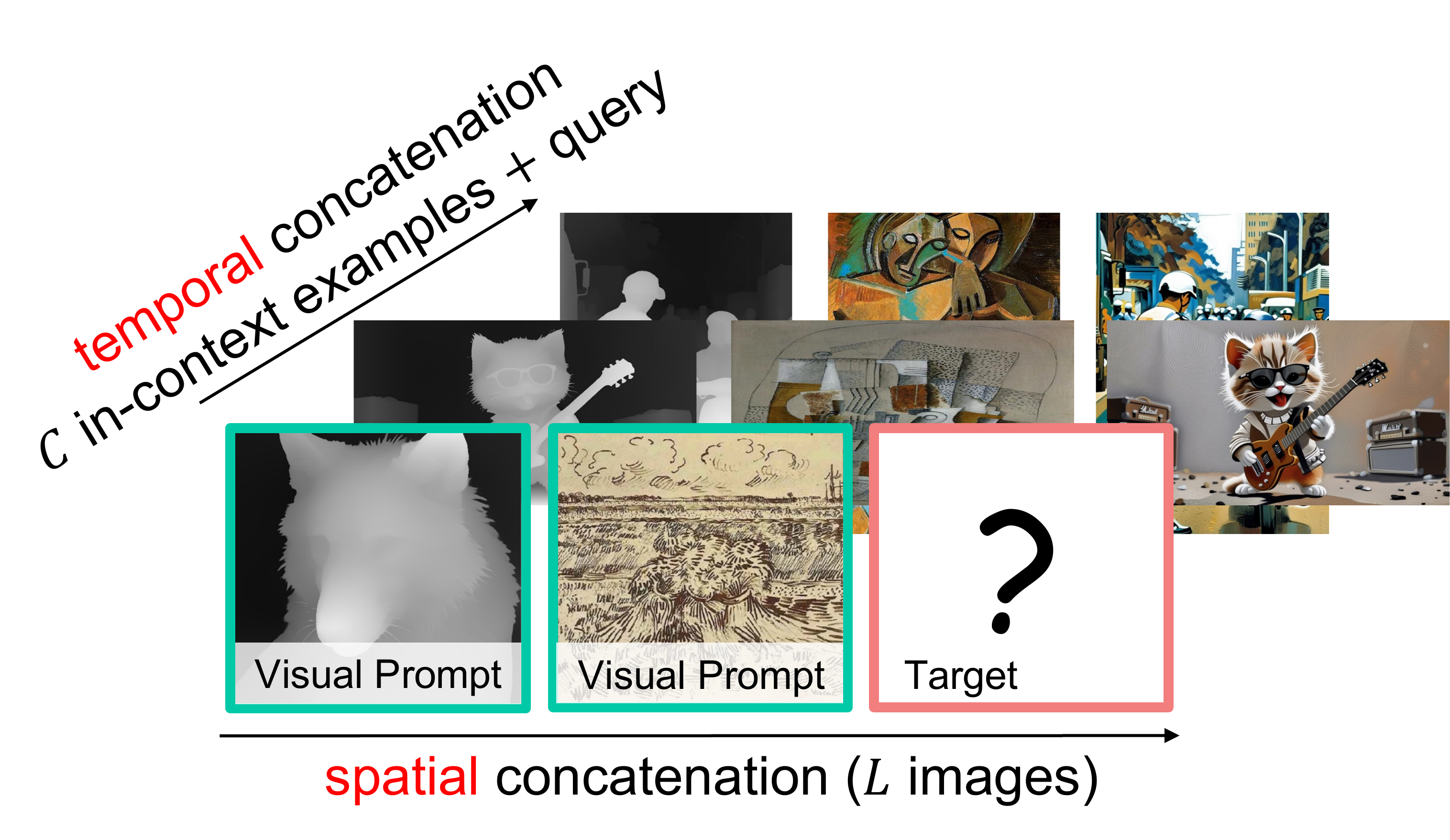}
	\end{overpic}
	\caption{{Concatenating images when applying position embeddings.} 
  The $L$ images within $C$ in-context examples and the query are first concatenated horizontally.
  Then, these concatenated rows are concatenated temporally to {{handle mismatched aspect ratios}}.
    }
    \label{fig:concatenation}
\end{figure}

\subsection{Unified Multi-task Framework} 
\label{sec:unified_framework}

Unlike previous visual in-context learning methods that primarily focus on scenarios with a single image condition and a single context~\cite{liu2023unifying, sun2023imagebrush}, in this work, we aim to construct a unified framework capable of handling varying numbers of conditions and contexts, allowing for flexible adaptation to diverse tasks.
For ease of description, 
we first assume all images processed by the model share the same size, $W \times H$, 
and we extend to the scenario with mismatched aspect ratios at the end of this section. 
{In this way, 
given $C$ in-context examples and the query, 
each containing $L$ images, 
all images can be concatenated into a complete grid-layout image with a 
size of $(L \times W, (C+1) \times H)$.} 
Then, the model can complete a task by infilling the target grids 
based on the surrounding context, akin to solving \emph{visual cloze} puzzles.
Therefore, we build our unified framework, \textbf{\ourmethod}, based on the general image infilling architecture capable of handling multiple resolutions.

Consistent with common diffusion-based infilling model designs, our model can be formulated as follows:
\begin{equation}
    \hat{X} = f(X \mid T, M), 
    \label{eq:single}
\end{equation}
where $X$ is the concatenated image, with the last grid left blank,
$T$ is the language instruction, 
$M$ is the mask condition, 
and $\hat{X}$ represents the infilled result. 
The mask $M$ is a binary matrix with the size of $(H\times(C+1), W\times L)$:
\begin{equation}
\label{eq:mask}
M(i, j) = \begin{cases}
    1 & \text{if } i \in \left[ H \times (C - 1), H \times C \right) \\
      & \text{and } j \in \left[ W \times (L - 1), W \times L \right), \\
    0 & \text{otherwise},
    \end{cases}
\end{equation}
where $M(i, j)=1$ indicates that the pixel will be masked and generated by the infilling model. 
{\eqref{eq:mask} masks the region in the last row and column, \ie, the target image. 
During training, 
we also randomly mask one of the first $L-1$ grids with a probability of 0.5, promoting reverse generation 
shown in \secref{sec:qualitative}.}
For the inference stage, we can crop $\hat{X}$ to obtain the target image easily.

\myPara{Aligned optimization objective.}
A key benefit of this design is that our \ourmethod~formulation shares a highly consistent objective with general image infilling models 
without architectural modifications or explicit input conditions. 
This consistency allows us to directly fine-tune advanced image infilling models using the newly constructed dataset while maximizing the utilization of 
the prior knowledge of foundation models.
In contrast, existing task-specific models often require introducing additional learnable modules~\cite{li2023photomaker,wei2024omniedit} or adapting to extra condition inputs~\cite{tan2024ominicontrol}, 
which may compromise the native capabilities of the model.

\myPara{Language instructions.}
Note that the design of language instruction is also necessary for \ourmethod~because 
it is responsible for defining the grid image layout, 
describing the caption of the image to be generated, 
and specifying the task intent when in-context examples are unavailable.
In our unified framework,
the instruction consists of three parts: 
(1) layout instruction, which describes the $(C+1)\times W$ layout of the grid image;
(2) task instruction, which specifies the task type; and
(3) content instruction, which describes the content of the target image.
The details about the instructions are available in \appref{app:instruction}. 
By restructuring the three components $X$, $T$, and $M$ in \eqref{eq:single}, 
we achieve a unified multi-task framework for image generation with the general image infilling paradigm
and support in-context learning.

\myPara{Positional embedding.}
In the preceding section, 
all images are concatenated into a grid-layout image and 
we can apply positional embedding (\ie, RoPE~\cite{ropepaper}) on this large image.
However, 
a potential limitation lies in composing a grid image from in-context examples with varying aspect ratios. 
To overcome this issue, 
we leverage the 3D-RoPE in Flux.1-Fill-dev to concatenate the query and in-context examples along the temporal dimension, 
as shown in \figref{fig:concatenation}, 
effectively overcoming this issue without introducing any noticeable performance degradation.

\subsection{Implementation Details}

We use FLUX.1-Fill-dev~\cite{flux2024} as our foundation model, considering its outstanding performance among open-source image infilling models.
{In this work, LoRA~\cite{hu2022lora} is chosen to fine-tune the model instead of fully fine-tuning it to reduce training costs and preserve the capabilities of the foundation model. 
The resulting LoRA can also be fused with other LoRAs in the community, enabling more widespread applications. 
Specifically, we set the rank of LoRA as $256$.}
The model is tuned for {20,000} iterations with an accumulated batch size of {64} on 8 $\times$ A100 GPUs. 
We employ the AdamW optimizer with a learning rate of $1e^{-4}$.
Following {FLUX.1-Fill-dev}, we incorporate the lognorm noise strategy with dynamic time shifting.
{During training, 
the number of in-context examples is set up to 2~(\ie, $C$ as defined in \secref{sec:unified_framework}), while $L$, the number of images involved in a task, varies between 2 and 4 in the Graph200K dataset. 
During inference, the number of in-context examples can be generalized to a larger number.}
{To balance computational efficiency, 
each image is resized to the area of $384 \times 384$ or $512 \times 512$ before concatenating them into a grid layout.}
High-resolution outputs can be obtained in practical applications through simple post-up-scaling techniques~\cite{meng2021sdedit}.

%% file: sec/5_experiment.tex
\section{Experiments}
\label{sec:experiments}

\subsection{Qualitative Analysis of In-context Learning}
\label{sec:qualitative}

This section presents a series of experiments demonstrating the effectiveness of in-context learning across different tasks, 
especially those unseen during training.
Based on our extensive experiments, 
we summarize four key findings 
that highlight the role of in-context learning.

\begin{figure}[t]
	\centering
	\begin{overpic}[width=1.0\linewidth]{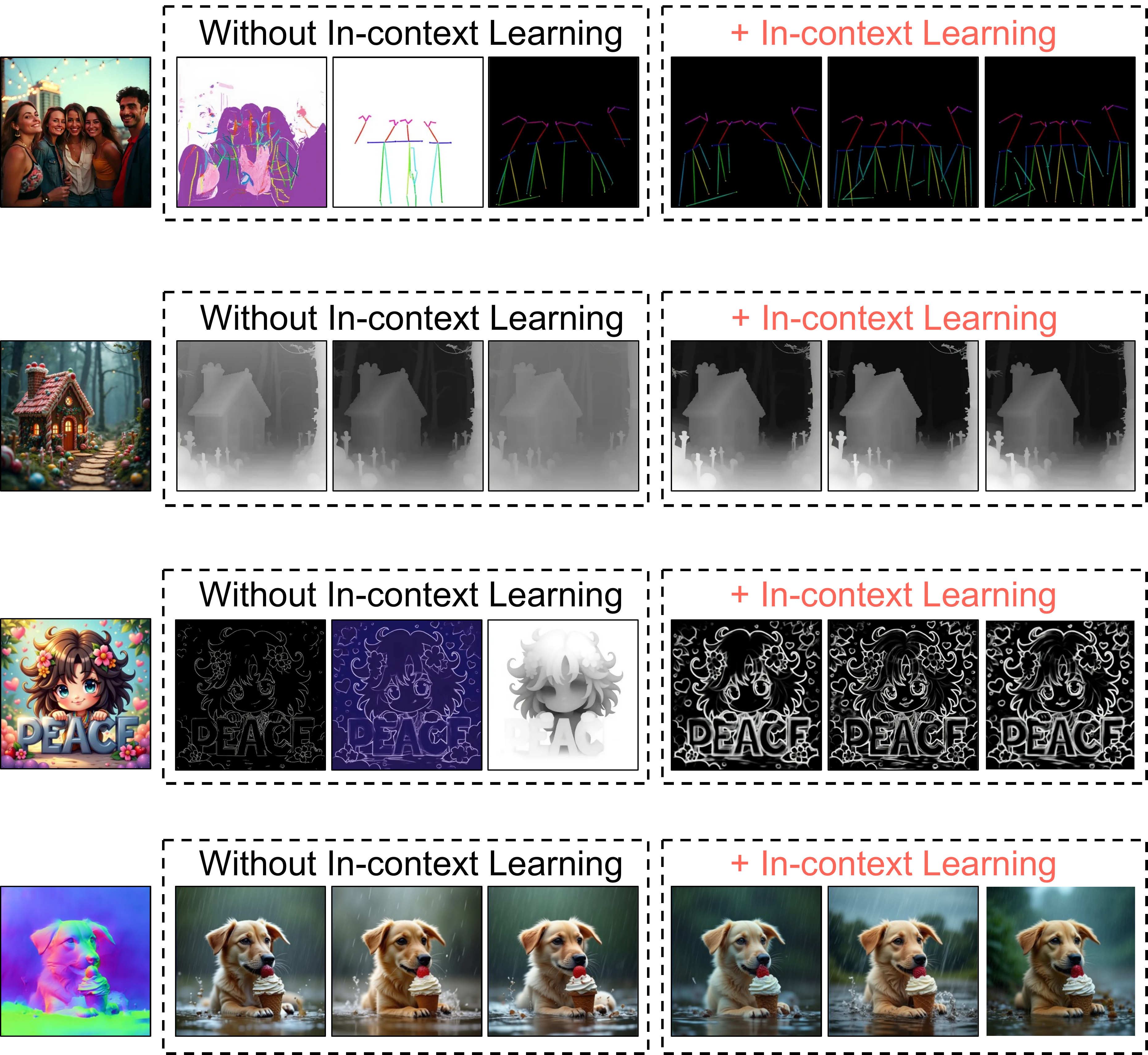}
        \put(42, 68.5){(a) Image to Pose}
        \put(41, 44){(b) Image to Depth}
        \put(42, 20){(c) Image to Edge}
        \put(39, -4){(d) Normal to Image}
    \end{overpic}
    \vspace{5pt}
	\caption{{In-context learning mitigates the 
    \textit{task ambiguity} in seen tasks.} 
    We show three results using different initial noises.}
    \label{fig:seen}
\end{figure}

\begin{tcolorbox}[colback=black!5!white,colframe=black!75!black,title=In-Context Learning Findings 1]
In-context learning can mitigate task confusion for seen tasks.
\end{tcolorbox}
\myPara{Task ambiguity on seen tasks.} 
The model occasionally experiences task confusion, 
failing to interpret the intended objective accurately, 
especially on dense prediction tasks. 
In-context learning effectively alleviates this issue by providing task-specific demonstrations. 
For example, 
in \figref{fig:seen}~(a) and (c), 
the model may produce noisy results without in-context examples in pose estimation and edge detection, 
while 
increasing the number of in-context examples enhances the performance and stability. 
In depth estimation shown in \figref{fig:seen}~(b), 
in-context examples also improve the accuracy 
when the model originally makes inaccurate estimates, especially in distant areas. 
Additionally, 
in some tasks like conditional generation, 
we note that the model can generate satisfactory results stably even without in-context examples, 
as shown in \figref{fig:seen}~(d).
However, 
the quantitative comparison in \tabref{tab:main_results} still shows that 
using in-context learning can further improve the accuracy of task completion.

\begin{figure*}[t]
	\centering
	\begin{overpic}[width=1.0\linewidth]{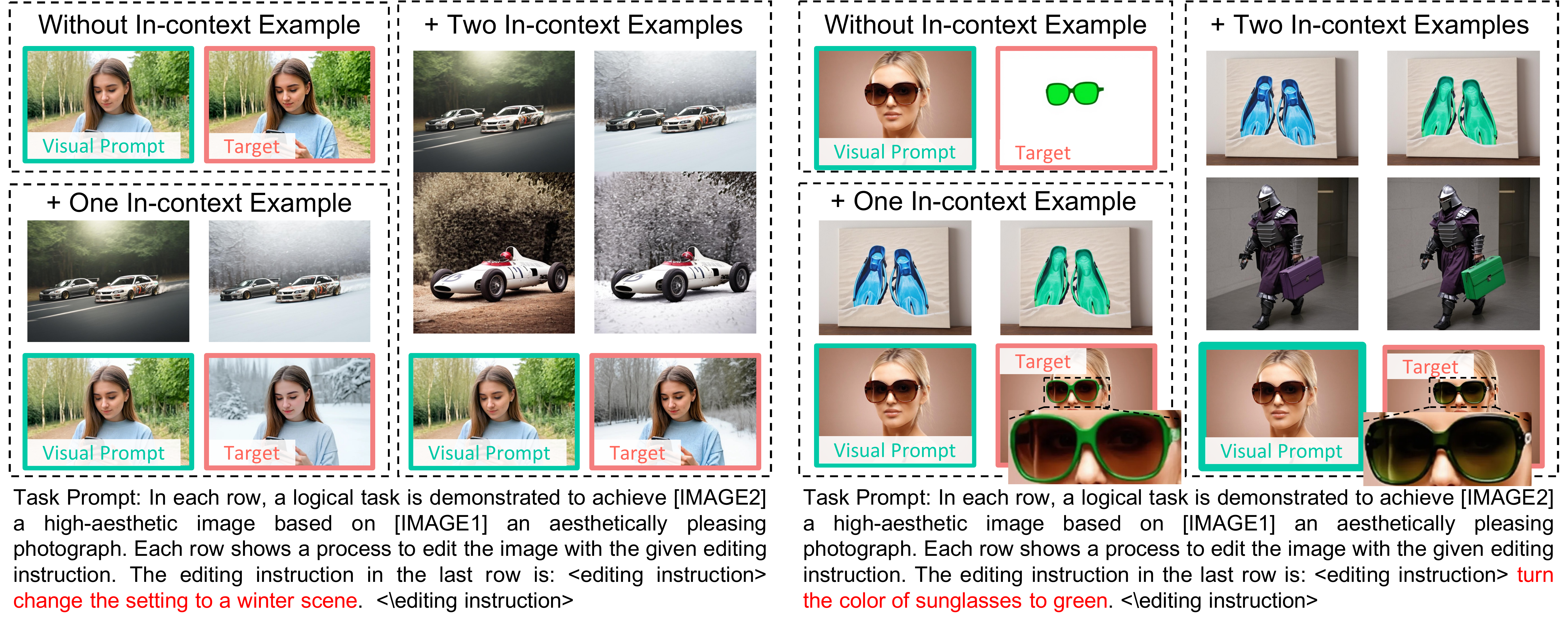}
    \end{overpic}
    \vspace{-20pt}
    \caption{\unseen~ 
    Although the image editing tasks seen by the model are only about object addition and object removal, 
    it can still generalize to other types of editing tasks, 
    such as environment modification (Left) and attribute transformation (Right), through in-context learning. 
    {More unseen tasks are shown in \figref{fig:unseen}.}} 
    \label{fig:unseen_edit}
    \vspace{-10pt}
\end{figure*}

\begin{figure}[t]
	\centering
	\begin{overpic}[width=1.0\linewidth]{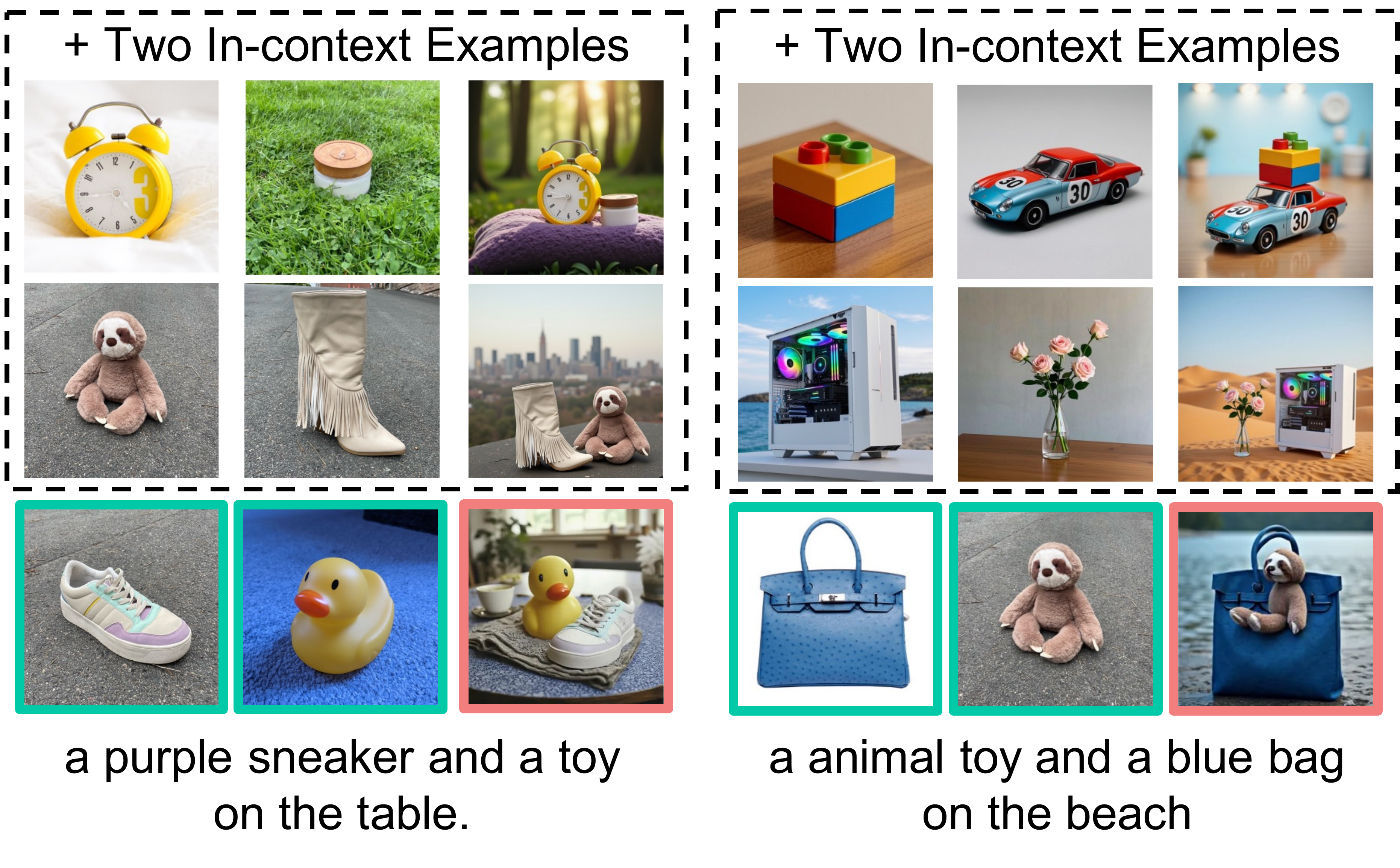}
	\end{overpic}
    \vspace{-20pt}
	\caption{\unseen~VisualCloze is capable of performing multi-subject driven generation~\cite{wu2025lesstomoregeneralizationunlockingcontrollability}, even though the model was only exposed to single subject-driven generation tasks during training. 
    Best viewed by zooming in.}
    \label{fig:multi_subject}
\end{figure}

\begin{tcolorbox}[colback=black!5!white,colframe=black!75!black,title=In-Context Learning Findings 2]
In-context learning supports generalization to unseen tasks, 
where providing more in-context examples could 
lead to more accurate generation.
\end{tcolorbox}
\myPara{Generalization on unseen tasks.} 
Beyond mitigating task confusion, in-context learning also enables the model to generalize to tasks unseen during training.
\figref{fig:unseen} has shown the model can successfully generate frontal faces from side-view images 
and transfer editing instructions~\cite{chen2025edittransferlearningimage} 
through in-context learning, 
even though they are not encountered during training. 
Here, we present additional examples of unseen tasks. 
{For instance, 
although the model is trained exclusively on image editing tasks involving object addition and removal, 
it still generalizes to other types of editing tasks, such as environment changes and attribute modifications, as shown in \figref{fig:unseen_edit}.
Furthermore, 
as demonstrated in \figref{fig:multi_subject}, 
the model, trained solely on single-subject generation, 
can generate images preserving identities of multiple subjects.} 
These results highlight that in-context learning is an effective guidance mechanism, 
enabling adaptation to novel tasks without retraining.

\begin{figure}[t]
	\centering
	\begin{overpic}[width=1.0\linewidth]{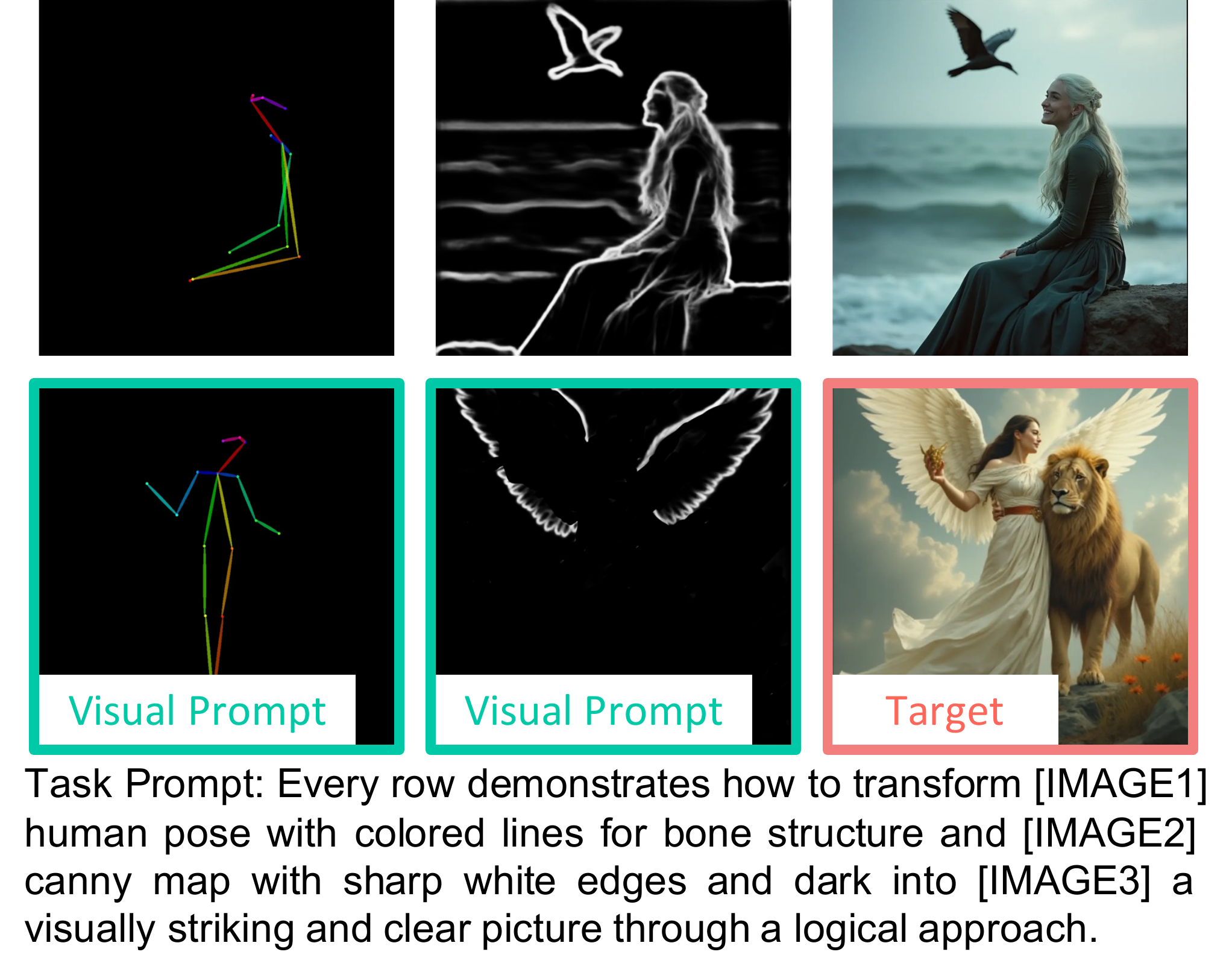}
    \end{overpic}
    \vspace{-15pt}
	\caption{\unseen~{\textit{Unseen combinations} of multiple tasks.} 
    For conditional generation, 
    we integrate multiple conditions achieve more precise control. 
    {More examples are shown in \figref{fig:combination}}.} 
    \label{fig:unseen_multi_condition}
\end{figure}

\begin{tcolorbox}[colback=black!5!white,colframe=black!75!black,title=In-Context Learning Findings 3]
{In-context learning enables task unification, an unseen strategy that consolidating sub-tasks into a single step and generating intermediate results. }
\end{tcolorbox}
\myPara{Multi-task consolidation.} 
Meanwhile, 
we also find that through in-context learning, 
we can consolidate multiple tasks into a single execution step, 
which can be viewed as another form of unseen task. 
\figref{fig:combination} has shown two examples, 
where we 1) merge conditional generation and relighting shown on the left and 
2) perform depth estimation, surface normal estimation, and edge detection simultaneously shown on the right. 
Similarly, \figref{fig:unseen_multi_condition} 
illustrates how we can combine multiple conditions for conditional generation to achieve finer control. 
For instance, 
generating a portrait based on keypoints provides only 
rough information about the location and body pose. 
In such cases, 
contour conditions can be used to control the attributes of 
other visual elements.

\begin{tcolorbox}[colback=black!5!white,colframe=black!75!black,title=In-Context Learning Findings 4]
Different in-context learning examples lead to varying effects, 
where examples that can better convey mission intent can achieve better and more stable generation. 
\end{tcolorbox}

\begin{figure}[t]
	\centering
	\begin{overpic}[width=1.0\linewidth]{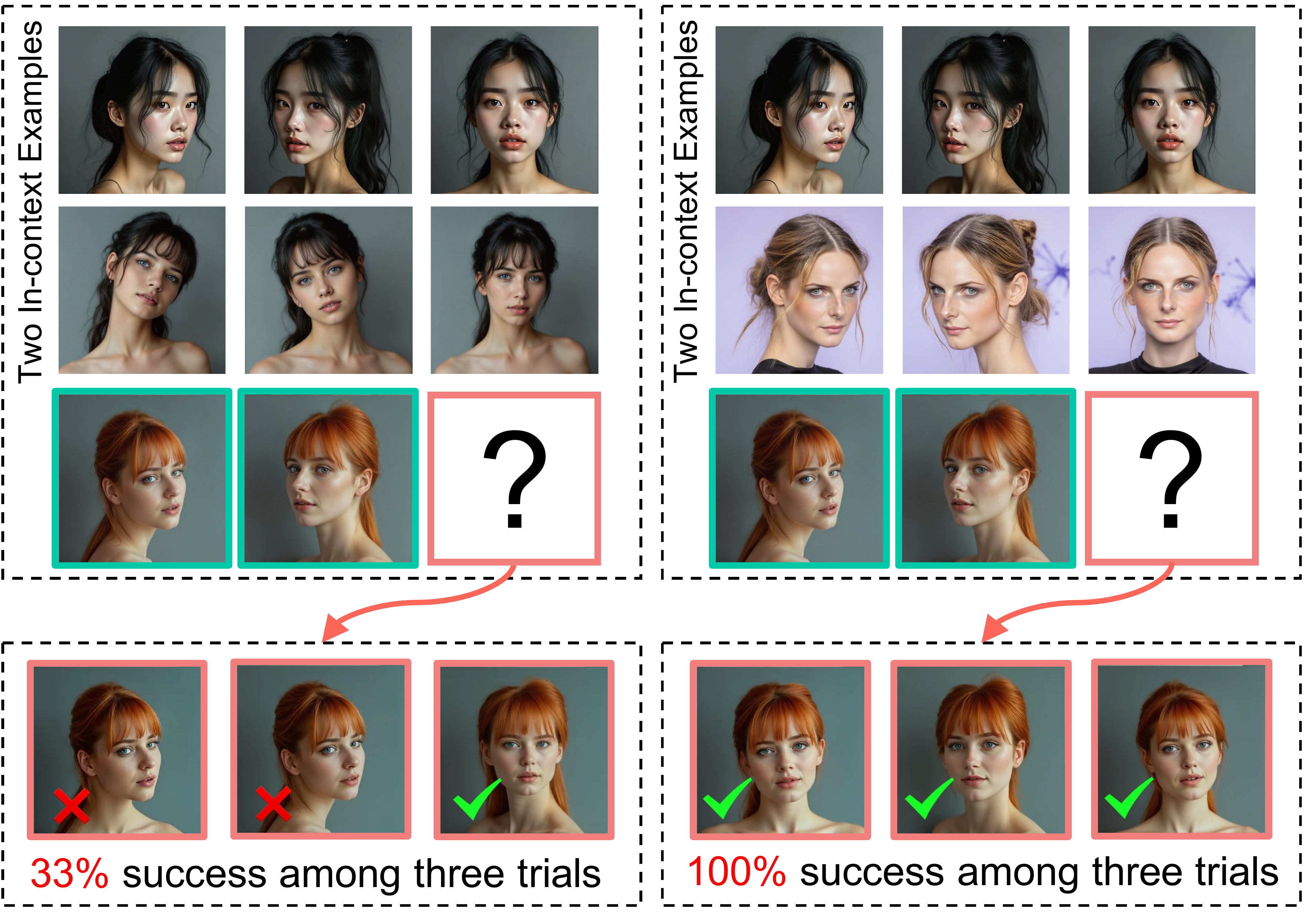}
    \end{overpic}
	\caption{{Illustration of the impact of different in-context examples on in-context learning}. 
    In the second example on the left, 
    the left and right faces are too biased towards the front, 
    so they do not show the core goal of the task intention.}
    \label{fig:different_incontext_examples}
\end{figure}

\myPara{Varying effects of different in-context examples.} 
Following prior works~\cite{rubin2022learningretrievepromptsincontext,10172590} on the prompt selection, 
we also find that different in-context examples 
could impact the generation quality. 
Specifically, 
it is crucial that 
in-context examples provide correct and strong guidance about 
the task intention. 
For example, 
as shown in \figref{fig:different_incontext_examples}~(left), 
when the side faces are more towards the front 
than in \figref{fig:different_incontext_examples}~(right), 
the success rate of correctly generating frontal faces 
has dropped dramatically.

\begin{figure*}[t]
	\centering
	\begin{overpic}[width=1.0\linewidth]{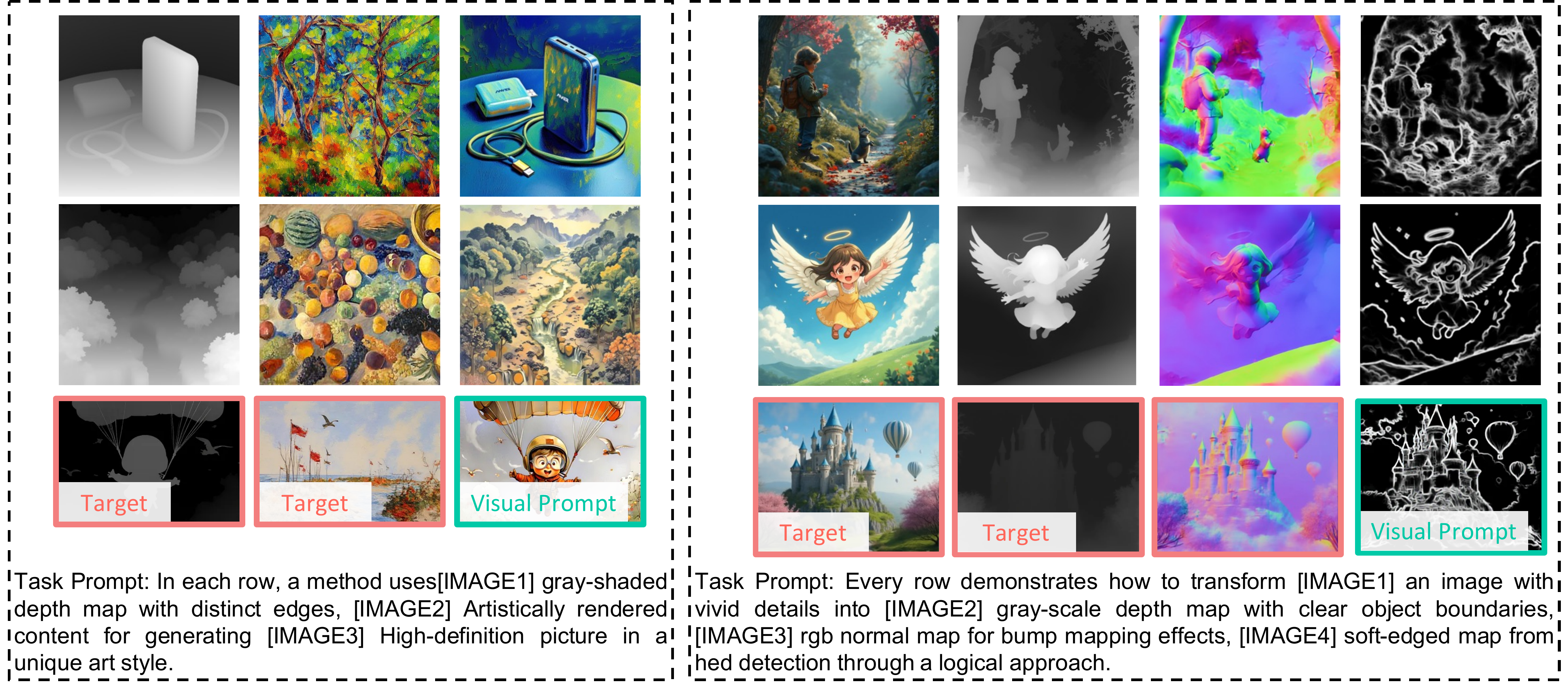}
        \put(1.3, 20.3){\rotatebox{90}{Two In-Context Examples}}
        \put(45.3, 20.3){\rotatebox{90}{Two In-Context Examples}}
	\end{overpic}
        \vspace{-20pt}
	\caption{\unseen~{Through in-context learning, 
    we can perform {\textit{reverse generation}}
    from targets to conditions.} 
    For example, 
    (a) decomposing the layout and style from a 
    stylized image and 
    (b) inferring the image, depth, and surface normal simultaneously 
    from an edge map, which is the reverse task of \figref{fig:combination} (Left).}
    \label{fig:reverse}
    \vspace{-10pt}
\end{figure*}

\begin{tcolorbox}[colback=black!5!white,colframe=black!75!black,title=In-Context Learning Findings 5]
In-context learning can guide bilateral generation, 
{even for the reverse process that is unseen during training.} 
\end{tcolorbox}

\input{table/tab1_control_result}

\myPara{Bilateral generation.}
In addition to generating the target from a set of given conditions, 
our model also shows the capability of reverse generation, 
\ie, inferring the underlying conditions from the target. 
{Although our model has randomly treated  
one condition image as the target when training as described in \secref{sec:unified_framework}, 
it can generalize to a more challenging and unseen setting during inference, \ie, inferring all conditional images from only the target image.} 
For instance, as illustrated in \figref{fig:reverse}~(left), 
the model can reverse-engineer both the original and 
the style reference images given a stylized image, 
demonstrating the ability to disentangle the content and style representations. 
Similarly, as shown in \figref{fig:reverse}~(right), 
the model can generate the corresponding real image, 
depth estimation, and surface normal estimation from an edge image, 
representing the inverse task of \figref{fig:combination}~(left). 
The ability to perform such reverse tasks highlights the 
flexibility and robustness in understanding complex relationships 
between different types of image representations. 

\input{table/subject_driven_result}

\input{table/style_transfer}

\subsection{Main Results}

We compare our method with 
universal generative models, including OmniGen~\cite{xiao2024omnigen} and OneDiffusion~\cite{le2024diffusiongenerate}, 
as well as  
specialized models, such as ControlNet~\cite{zhang2023adding} and OminiControl~\cite{tan2024ominicontrol}. 
The details of the evaluation metrics are provided in \appref{app:metric}. 
Additionally, we fine-tune FLUX.1-dev~\cite{flux2024} using the same settings as FLUX.1-Fill-dev for comparison 
and refer to the tuned models as Ours$_{\rm dev}$ and Ours$_{\rm fill}$. 
The details of Ours$_{\rm dev}$ are shown in \appref{app:flux_dev}. 

For conditional generation and image restoration, 
we evaluate the models based on three criteria, \ie, controllability, visual quality, and text consistency, 
following the evaluation approach of OminiControl~\cite{tan2024ominicontrol}.
As shown in \tabref{tab:main_results}, 
our framework demonstrates comparable controllability to existing universal methods 
while achieving superior visual quality and text consistency. 
Compared to specialized methods, 
our model performs on par with the best results 
and even outperforms them on the depth-to-image. 

In the style transfer task, 
we measure text consistency and style alignment using the CLIP~\cite{radford2021learning} model. 
As reported in \tabref{tab:style}, 
our method outperforms OmniGen~\cite{xiao2024omnigen} by 2\% and 3\% in text alignment and style consistency, respectively. 
Even when compared with InstantStyle-Plus~\cite{Zhang_2023_inst}, a specialized model, 
we achieve a 2\% improvement in text consistency, with only a slight decrease in style alignment. 

Furthermore, we evaluate the models on subject-driven image generation and report semantic alignment 
using the DINOv2~\cite{oquab2023dinov2}, CLIP-I~\cite{radford2021learning}, and CLIP-T~\cite{radford2021learning} scores. 
Across all these metrics, our method consistently delivers improvements, 
as shown in \tabref{tab:subject}. 
For example, compared to the specialized model OminiControl~\cite{tan2024ominicontrol}, 
we achieve improvements of 7.15\%, 1.66\%, and 1.48\% in these three scores.

\myPara{Advantages of the infilling model.}
Our method (Ours$_{\rm fill}$) is built on FLUX.1-Fill-dev~\cite{flux2024}, 
which shares the same objective as our unified image generation framework. 
To verify its effectiveness, 
we also fine-tune Fill.1-dev~\cite{flux2024} (Ours$_{\rm dev}$) using identical settings. 
Unlike Ours$_{\rm fill}$, which requires no modifications, 
Ours$_{\rm dev}$ necessitates model adaptations for universal image generation, 
as shown in \appref{app:flux_dev}. 
Despite its simplicity, Ours$_{\rm fill}$ achieves superior performance across multiple tasks.

\begin{figure}[t]
	\centering
	\begin{overpic}[width=1.0\linewidth]{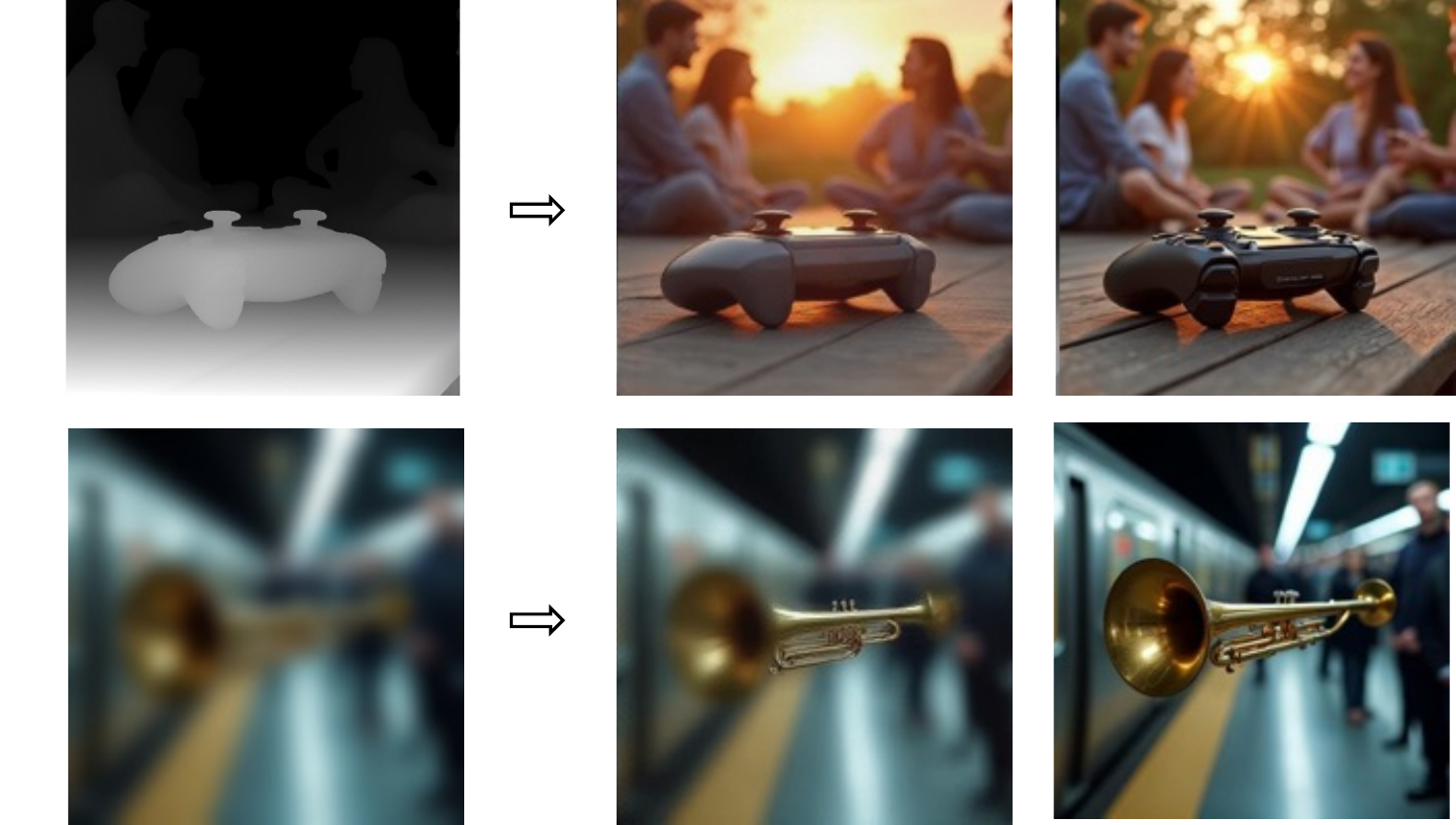}
        \put(10.0, -3.5){Condition}
        \put(51.8, -3.5){Ours$_{dev}$}
        \put(80.5, -3.5){Ours$_{fill}$}
        \put(0.0, 29){\rotatebox{90}{Depth to Image}}
        \put(0.0, 3.9){\rotatebox{90}{Debluring}}
	\end{overpic}
    \vspace{3pt}
	\caption{Comparison between Flux.1-dev~(Ours$_{dev}$) 
    and Flux.1-Fill-dev~(Ours$_{fill}$).}
    \label{fig:quality}
\end{figure}

As shown in \tabref{tab:main_results}, 
Ours$_{\rm dev}$ achieves a higher F1 score than Ours$_{\rm fill}$ in the canny-to-image generation. 
However, in other tasks, Ours$_{\rm fill}$ demonstrates a significant advantage.
For instance, in the depth-to-image generation, 
Ours$_{\rm fill}$ reduces RMSE from 25.06 to 10.31. 
In the deblurring task, Ours$_{\rm fill}$ achieves superior quality by lowering RMSE while maintaining a higher SSIM.
In subject-driven image generation, 
\tabref{tab:subject} shows that Ours$_{\rm fill}$ consistently outperforms Ours$_{\rm dev}$. 
Additionally, in semantic-invariant style transfer, 
Ours$_{\rm fill}$ delivers comparable performance to Ours$_{\rm dev}$, as shown in \tabref{tab:style}.

\figref{fig:quality} presents a visual comparison, where Ours$_{\rm fill}$ demonstrates clear advantages over Ours$_{\rm dev}$. 
Notably, in the depth-to-image generation, images produced by Ours$_{\rm dev}$ 
frequently exhibit diagonal streak artifacts, which significantly degrade visual fidelity.
Considering the advantages in performance, visual quality, and architectural efficiency, 
Ours$_{\rm fill}$ stands out as the superior model.

\myPara{Quantitative comparison on in-context learning.}
{Here, we further analyze the impact of in-context learning on seen tasks.
\tabref{tab:main_results} demonstrates the impact of in-context learning on 
different image generation tasks. 
Under the canny condition, 
our method without in-context examples achieves an FID of 30.60, 
which improves to 31.15 with two in-context examples. 
When conditioned on depth, the RMSE decreases from 10.31 to 9.68 as the number of in-context examples increases, 
indicating enhanced structural consistency. 
Similarly, in the deblurring task, RMSE decreases from 26.53 to 25.57, 
reflecting improved fidelity to the original content.
These results highlight in-context learning as an effective guidance mechanism, 
enabling the model to better align with the task intent.}

%% file: table/tab1_control_result.tex
\newcommand{\best}[1]{\textbf{#1}}

\begin{table*}
  [!t] \small
  \centering
  \setlength{\tabcolsep}{1.1mm}    
  \begin{tabular}{clcccccccc}
    \toprule
    \multirow{2}{*}{Condition}               & \multirow{2}{*}{Method}  &  \multirow{2}{*}{Context}  &  \multicolumn{2}{c}{Controllability}                          & \multicolumn{4}{c}{Quality} & Text Consistency                     \\
    \cmidrule(lr){4-5} \cmidrule(lr){6-9} \cmidrule(lr){10-10} 
                                             &               &               & $\text{F1} \uparrow$ & $\text{RMSE} \downarrow$ & $\text{FID~\cite{heusel2017gans}}\downarrow$               & $\text{SSIM}\uparrow$               & $\text{MAN-IQA~\cite{yang2022maniqa}}\uparrow$            & $\text{MUSIQ~\cite{ke2021musiq}}\uparrow$               & $\text{CLIP-Score~\cite{radford2021learning}}\uparrow$          \\
     \midrule \multirow{8}{*}{Canny} & \color{gray} ControlNet~\cite{zhang2023adding} & & \color{gray} {0.13} & \color{gray} - & \color{gray} 46.06  & \color{gray} 0.34 & \color{gray} 0.31 & \color{gray} 45.45 & \color{gray} 34.10 \\
                                             & \color{gray} OminiControl~\cite{tan2024ominicontrol}  &  & \color{gray} 0.47 & \color{gray} - & \color{gray} 29.58 & \color{gray} 0.61 & \color{gray} 0.44 & \color{gray} 61.40 & \color{gray} 34.40 \\
                                             & OneDiffusion~\cite{le2024diffusiongenerate} & & \underline{0.39} & - & 32.76 & 0.55 & 0.46 & 59.99 & \underline{34.99} \\
                                             & OmniGen~\cite{xiao2024omnigen}  &   & \textbf{0.43} & - &  51.58  &  {0.47} & 0.47 & 62.66 & 33.66  \\
                                             & Ours$_{\rm dev}$  & 0 & \underline{0.39}  & - &   \textbf{30.36} & \textbf{0.61} & \underline{0.48} & 61.13 & \textbf{35.03}   \\
                                             & Ours$_{\rm fill}$  & 0 & 0.35 & - &    \underline{30.60}  & 0.55 & \textbf{0.49} & \textbf{64.39} & 34.98 \\
                                             & Ours$_{\rm fill}$  &  1   & 0.36 & - & 31.34 & 0.55 & \textbf{0.49} & \underline{64.12} & 34.96  \\
                                             & Ours$_{\rm fill}$  &  2   & 0.36 & - & 31.15 &  \underline{0.56} & \textbf{0.49} & 64.08 & 34.85  \\
    \midrule \multirow{8}{*}{Depth}          & \color{gray} ControlNet~\cite{zhang2023adding} & & \color{gray} - & \color{gray} 23.70 & \color{gray} 36.83 & \color{gray} 0.41 & \color{gray} 0.44 & \color{gray} 60.17 & \color{gray} 34.49 \\
                                             & \color{gray} OminiControl~\cite{tan2024ominicontrol} & & \color{gray} - & \color{gray} 21.44 & \color{gray} 36.23 & \color{gray} 0.52 & \color{gray} 0.44 & \color{gray} 60.18 & \color{gray} 34.08 \\
                                             & OneDiffusion~\cite{le2024diffusiongenerate} & & - & 10.35 & 39.03 & 0.49 & \textbf{0.49} & 60.49 & 34.71   \\
                                             & OmniGen~\cite{xiao2024omnigen}  &   & - &    15.07    &  86.08   &    0.26  & \textbf{0.49} & \textbf{64.90} & 29.72   \\
                                             & Ours$_{\rm dev}$ & 0 & - & 25.06 &   42.14  & \underline{0.53} & 0.46 & 58.95 & 34.80  \\
                                             & Ours$_{\rm fill}$  & 0 & - & 10.31  & \textbf{33.88}  &   \textbf{0.54} & \underline{0.48} & \underline{64.85} & \textbf{35.10}    \\ 
                                             & Ours$_{\rm fill}$  &  1 & - & \underline{9.91}   & \underline{34.44}  & \textbf{0.54} & \textbf{0.49} & 64.32 & \underline{34.95}   \\
                                             & Ours$_{\rm fill}$  &  2  & - & \textbf{9.68} & {34.88} & \textbf{0.54} & \underline{0.48} & 64.29 & 34.89 \\
    \midrule \multirow{8}{*}{Deblur}         & \color{gray} ControlNet~\cite{zhang2023adding}  & & \color{gray} - & \color{gray} 37.82 & \color{gray} 53.28 & \color{gray} 0.49 & \color{gray} 0.45 & \color{gray} 61.92 & \color{gray} 33.80 \\
                                             & \color{gray} OminiControl~\cite{tan2024ominicontrol} & & \color{gray} - & \color{gray} 19.70 & \color{gray} 26.17 & \color{gray} 0.85 & \color{gray} 0.45 & \color{gray} 60.70 & \color{gray} 34.53 \\
                                             & OneDiffusion~\cite{le2024diffusiongenerate}    & & - & - & - & - & - & - & - \\
                                             & OmniGen~\cite{xiao2024omnigen} & & - & - & - & - & - & - & - \\
                                             & Ours$_{\rm dev}$  &  0  & - & \textbf{25.03} & 56.76 & \underline{0.74} & 0.38 & 46.68 & 33.52 \\
                                             & Ours$_{\rm fill}$  &  0 & - & 26.53 & 40.59 & \underline{0.74} & \underline{0.46} & 59.62 & \underline{34.56} \\
                                             & Ours$_{\rm fill}$  &  1 & - & 25.87 & \underline{36.93} & \textbf{0.76} & \textbf{0.48} & \underline{61.58} & \textbf{34.82} \\
                                             & Ours$_{\rm fill}$  &  2 & - & \underline{25.57} & \textbf{36.28} & \textbf{0.76} & \textbf{0.48} & \textbf{61.77} & \textbf{34.82} \\
    \bottomrule
  \end{tabular}
  \caption{Quantitative comparison on conditioning generation and image restoration. 
  The methods that train a specialist for each task are marked as {\color{gray}{gray color}}. 
  Except for these methods, 
  the best method is bolded, and the second best method is \underline{underlined}.}
  \label{tab:main_results}
  \vspace{-10pt}
\end{table*}

%% file: table/subject_driven_result.tex
\begin{table}[t]
    \centering   
    \setlength{\tabcolsep}{0.7mm}
    \begin{tabular}{lcccc}
        \toprule
        {Method} &  {Context} & {DINOv2} & {CLIP-I} & {CLIP-T} \\
        \midrule
        \color{gray} OminiControl~\cite{tan2024ominicontrol} & & \color{gray} 73.17 & \color{gray} 87.70 & \color{gray} 33.53  \\
        OneDiffusion~\cite{le2024diffusiongenerate} & & 73.88 & 86.91 & 34.85 \\
        OmniGen~\cite{xiao2024omnigen} &  & 67.73 & 83.43 & 34.53  \\
        \midrule
        Ours$_{\rm dev}$  & 0 & 78.05 & 87.68  & \underline{35.06} \\
        Ours$_{\rm fill}$ & 0 & \textbf{80.41} & \textbf{89.63} & \textbf{35.16} \\
        Ours$_{\rm fill}$ & 1 & 79.33 & 89.22 & 35.02 \\
        Ours$_{\rm fill}$ & 2 & \underline{80.32} & \underline{89.36} & 35.01 \\
        \bottomrule
    \end{tabular}
    \caption{Quantitative comparison for subject-driven image generation. We report clip
    scores on text alignment and style consistency. 
    Specialists are shaded in {\color{gray}{gray}}.
    Among the remaining methods, the best is emphasized in bold, while the second best is \underline{underlined}.}
    \label{tab:subject}
\end{table}

%% file: table/style_transfer.tex
\begin{table}[t]
  \centering   
  \setlength{\tabcolsep}{6.7mm}    
  \begin{tabular}{lccccccccccccc}
      \toprule
      & text$\uparrow$ & image$\uparrow$ \\
      \midrule
      \color{gray} InstantStyle~\cite{wang2024instantstyle} & \color{gray} 0.27 & \color{gray} 0.60 \\
      OmniGen~\cite{xiao2024omnigen} & 0.27 & 0.52 \\
      Ours$_{\rm dev}$ & \textbf{0.30} & \underline{0.53} \\
      Ours$_{\rm fill}$ & \underline{0.29} & \textbf{0.55} \\
      \bottomrule
      \end{tabular}
  \caption{Quantitative comparison for style transfer. 
  We report CLIP scores on text alignment and style consistency. 
  The specialists are indicated in {\color{gray}{gray}}.
  Among the others, 
  the top-performing one is highlighted in bold, 
  and the second best is \underline{underlined}.}
  \label{tab:style}
  \vspace{-10pt}
\end{table}

%% file: sec/6_limitation.tex
\section{Limitations}
\label{sec:limitations}

While our model demonstrates strong stability across most in-domain tasks, it still exhibits some instability in specific tasks, such as object removal. 
This limitation suggests that the performance is sensitive to certain task characteristics. 
Additionally, 
the stability of the model on unseen tasks is still insufficient. 
Apart from the difficulty of the task and the difference with seen tasks, 
ambiguous in-context examples may also lead to less stable results, 
as discussed in \secref{sec:qualitative}.

%% file: sec/7_conclusion.tex
\vspace{-6pt}
\section{Conclusion}
\label{sec:conclusion}

In this work, we propose \ourmethod, a universal image generation framework 
that addresses key challenges in existing methods, 
including generalizable instruction design, appropriate task distributions, and unified architectural design. 
Rather than relying solely on language-based instructions to convey task intent, 
we re-propose visual in-context learning, 
enabling the model to learn tasks from a few demonstrations. 
This approach improves generalization to unseen tasks and reduces task ambiguity. 
To overcome the sparsity of visual task distributions, which limits the learning of transferable knowledge, 
we construct \ourdataset, a graph-structured dataset that establishes interrelated tasks. 
In this compact task space, the model is promoted to learn transferable representations and improve adaptability. 
Meanwhile, we identify the consistent  objective 
between image infilling and our universal generation formulation , 
allowing us to seamlessly adapt general-purpose infilling models for universal generation without architectural modifications. 
Experimental results show that our approach supports 
a diverse set of in-domain tasks using in-context learning 
while demonstrating strong generalization to unseen tasks.

\section{Acknowledgments}
This work is supported by the National Natural Science Foundation of China (Grant No. 62225604 and 62206272), and the Shenzhen Science and Technology Program (JCYJ20240813114237048).

%% file: sec/8_appendix.tex
\clearpage

\begin{appendices}

\begin{figure*}[t]
\centering
\begin{overpic}[width=0.95\linewidth]{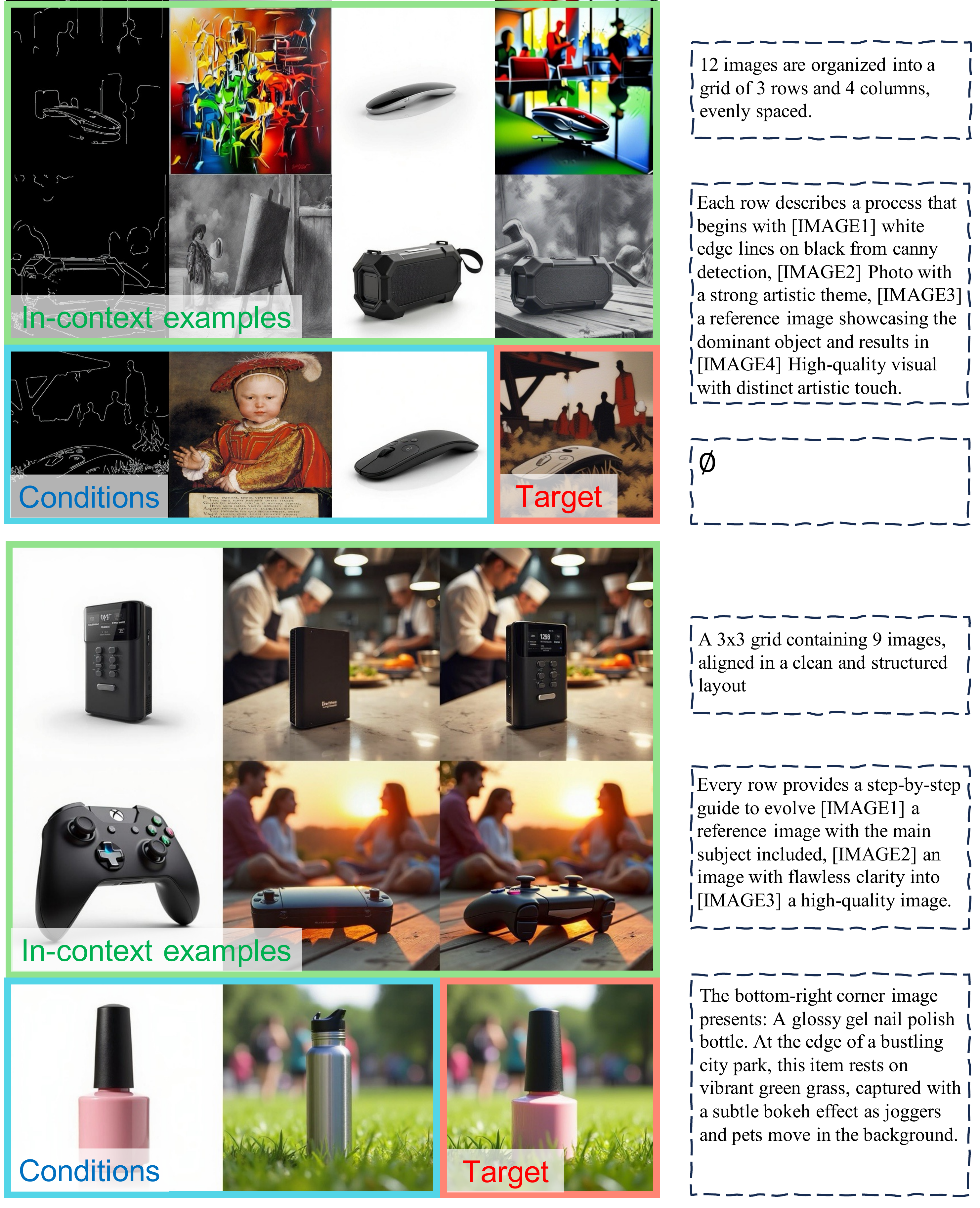}
  \put(57.5, 50.5){Layout instruction:}
  \put(57.5, 38.0){Task instruction:}
  \put(57.5, 20.8){Content instruction:}
  \put(57.5, 97.8){Layout instruction:}
  \put(57.5, 86.0){Task instruction:}
  \put(57.5, 64.6){Content instruction:}
  \put(19, -1){(a) Concatenated images}
  \put(60, -1){(b) Language instructions}
\end{overpic}
\caption{Examples of language instructions that contain prompts about the layout of the concatenated image, 
task intent, and content of the target image.}
  \label{fig:instruction1}
\end{figure*}

\begin{figure*}[t]
	\centering
	\begin{overpic}[width=1.0\linewidth]{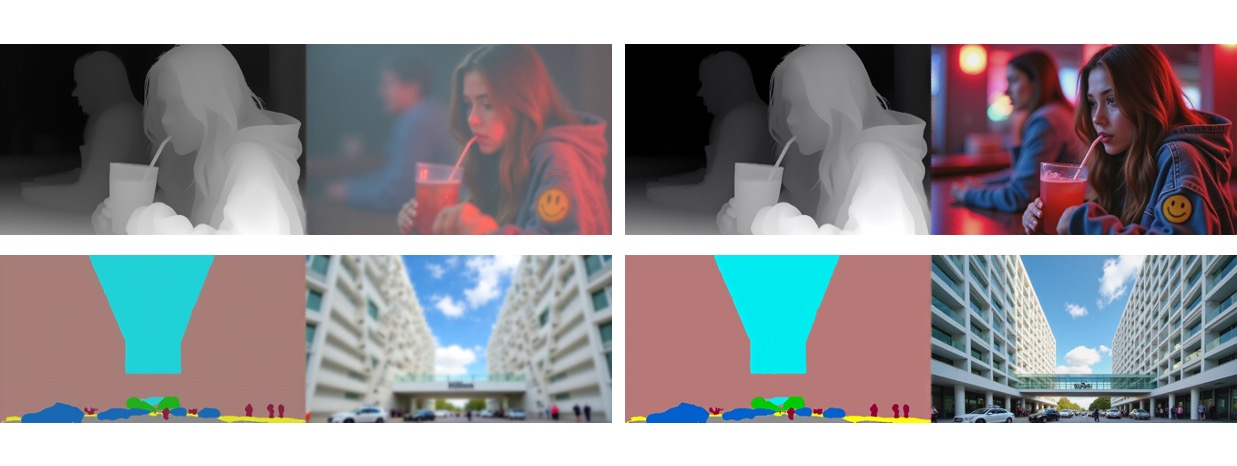}
    \put(14.0, 0){(a) separate mean and shift}
    \put(65, 0){(b) unified mean and shift}
    \put(8.0, 35){Condition}
    \put(34.0, 35){Target}
    \put(58.5, 35){Condition}
    \put(85.0, 35){Target}
	\end{overpic}
	\caption{Effects of separate mean and shift in fine-tuning FLUX.1-dev.}
    \label{fig:separate_adaLN}
    \vspace{-10pt}
\end{figure*}

\section{Instruction Format}
\label{app:instruction}

In our unified framework,
the instruction consists of three parts: (1) layout instruction, which describes the layout of the grid image;
(2) task instruction, which specifies the task type; and
(3) content instruction, which describes the content of the target image. 
\figref{fig:instruction1} illustrates the instructions 
for concept fusion of style, subject, and layout~(\figref{fig:instruction1} upper) and 
image editing with reference~(\figref{fig:instruction1} bottom). 
The content instruction is omitted for some tasks that provide strong visual cues in conditions, like style transfer.

\section{Fine-tuning FLUX.1-dev Model}
\label{app:flux_dev}
Apart from FLUX.1-Fill-dev, 
we also adapt our method to FLUX.1-dev~\cite{flux2024}, 
a common text-to-image generative model. 
Unlike the infilling model that shares a consistent objective with 
universal image generation, 
FLUX.1-dev requires customized modifications to 
process clean condition images and noise target images. 
Specifically, 
after concatenating images in a grid layout like the infilling model, 
we always keep the region corresponding to the conditions as clean latent embeddings 
throughout the sampling process. 
This strategy requires modifications in image sampling 
because FLUX.1-Fill-dev takes noise latent embeddings as input. 
Moreover, 
for the adaLN-Zero block~\cite{Peebles_2023_ICCV}, 
it is critical 
to calculate the separate mean and shift parameters for the regions of clean conditions and noise target 
by feeding $T=0$ and $T=t$ into the adaLN-Zero, respectively. 
$t$ indicates the timestep in each sampling step and gradually increases from 0 to 1 along the sampling process.
This strategy aligns with the pre-training domain of FLUX.1-dev, 
where different noise levels correspond to different mean and shift. 
As shown in \figref{fig:separate_adaLN}, 
this strategy ensures the visual fidelity. 

\section{Evaluation Metrics}
\label{app:metric}

\subsection{Conditioning Generation}
We assess the models from controllability, quality, and text consistency to evaluate image generation quality in conditioning generation and image restoration tasks. 

\myPara{Controllability.} 
For conditional image generation, 
we measure the difference between 
the input conditions and those extracted from generated images. 
Specifically, 
we calculate the F1 Score for the cany-to-image task and 
RMSE for the depth-to-image task. 
Additionally, 
for debluring, 
we measure the RMSE between 
original and restored images. 

\myPara{Generation quality.}
We measure the Generation quality using FID~\cite{heusel2017gans}, SSIM, 
MAN-IQA~\cite{yang2022maniqa}, 
and MAN-IQA~\cite{yang2022maniqa}.
FID~\cite{heusel2017gans} measures the similarity between generated and real image feature distributions. 
SSIM evaluates perceptual quality by comparing luminance, contrast, and structural patterns between images. 
It calculates local patch statistics and combines them into a composite score ranging from $-1$ to $1$, with higher values indicating better structural preservation.
MANIQA~\cite{yang2022maniqa} and MUSIQ~\cite{ke2021musiq} leverage neural networks 
to predict image quality scores. 

\paragraph{Text consistency.}
Leveraging the powerful multi-modal capability of CLIP~\cite{radford2021learning}, 
we also measure the semantic alignment between generated images and text prompts, 
which reflects how the model follows instructions. 

\subsection{Subject Driven Generation}
Following DreamBooth~\cite{ruiz2023dreambooth} and BLIP-Diffusion~\cite{li2023blip}, 
we measure DINOv2~\cite{oquab2023dinov2}, CLIP-I~\cite{radford2021learning}, 
and CLIP-T scores for the comparison of subject-driven image generation. 
DINOv2~\cite{oquab2023dinov2} and CLIP-I scores measure the alignment between the reference subject and generated images  
through cosine similarity and CLIP score, respectively. 
CLIP-T measures the alignment between the generated image and the corresponding text prompt. 

\subsection{Style Transfer}
Following StyleDrop~\cite{sohn2023styledrop}, 
we assess the performance of style transfer 
according to text consistency and style alignment. 
For text alignment, 
we measure the cosine similarity between embeddings of generated images and text prompts, 
where the embeddings are extracted by CLIP~\cite{radford2021learning}. 
Regarding style consistency, 
we measure the cosine similarity between embeddings of generated images and style reference. 
Note that these two metrics should be considered together 
because the style consistency will reach 1.0 if the model collapses, 
where the model completely copies style reference as a composite image and 
ignores text instructions. 

\end{appendices}